\pdfoutput=1
\documentclass[11pt,a4paper]{article}
\usepackage[hyperref]{acl2021}
\usepackage{times}
\usepackage{latexsym}

\usepackage{amsfonts}
\newcommand{\finetunevar}{\mathrm{FineVar}}
\newcommand{\ckptvar}{\mathrm{CkptVar}}
\newcommand{\pretrainvar}{\mathrm{PretVar}}
\newcommand{\biassquared}{\mathrm{Bias^{2}}}
\newcommand{\bias}{\mathrm{Bias}}
\newcommand{\difff}{\text{Diff}_{\mathrm{FTune}}}
\newcommand{\diffp}{\text{Diff}_{\mathrm{PTrain}}}

\newcommand{\acc}{\mathrm{Acc}}
\newcommand{\avgstd}{\text{Std}_{\mathrm{all}}}

\newcommand{\decay}{\mathrm{Decay}}
\usepackage[export]{adjustbox}
\usepackage[caption=false]{subfig}
\usepackage[ruled,vlined, noend]{algorithm2e}

\newcommand{\minisize}{\textsc{mini} }
\newcommand{\smallsize}{\textsc{small} }
\newcommand{\mediumsize}{\textsc{medium} }
\newcommand{\basesize}{\textsc{base} }
\newcommand{\largesize}{\textsc{large} }
\usepackage{subfig}
\usepackage{graphicx}
\usepackage{amssymb}
\usepackage{amsmath}
\usepackage{ntheorem}
\newtheorem{theorem}{Theorem}
\newtheorem{lemma}{Lemma}
\newcommand{\Var}{\operatorname{Var}}

\usepackage{microtype}
\usepackage{siunitx}
\aclfinalcopy 

\title{Are Larger Pretrained Language Models Uniformly Better? Comparing Performance at the Instance Level}

\author{
Ruiqi Zhong \quad Dhruba Ghosh \quad Dan Klein \quad Jacob Steinhardt \\
Computer Science Division, University of California, Berkeley \\
\{ruiqi-zhong, djghosh13, klein, jsteinhardt\}@berkeley.edu
}

\date{}

\begin{document}
\maketitle
\begin{abstract}
Larger language models have higher accuracy on average, but are they better on every single instance (datapoint)? 
Some work suggests larger models have higher out-of-distribution robustness, while other work suggests they have lower accuracy on rare subgroups. 
To understand these differences, we investigate these models at the level of individual instances. 
However, one major challenge is that individual predictions are highly sensitive to noise in the randomness in training. We develop statistically rigorous methods to address this, and after accounting for pretraining and finetuning noise, we find that our BERT-\largesize is worse than BERT-\minisize on at least 1$-$4\% of instances across MNLI, SST-2, and QQP, compared to the overall accuracy improvement of 2$-$10\%. We also find that finetuning noise increases with model size, and that instance-level accuracy has momentum: improvement from BERT-\minisize to BERT-\mediumsize correlates with improvement from BERT-\mediumsize to BERT-\largesize.
Our findings suggest that instance-level predictions provide a rich source of information; we therefore recommend that researchers supplement model weights with model predictions.
\end{abstract}

\section{Introduction}
Historically, large deep learning models \cite{peters-etal-2018-deep, devlin-etal-2019-bert, lewis-etal-2020-bart, raffel2019exploring} have improved the state of the art on a wide range of tasks and leaderboards \cite{schwartz-etal-2014-machine, rajpurkar-etal-2016-squad, wang-etal-2018-glue},
and empirical scaling laws predict that larger models will continue to increase performance \cite{kaplan2020scaling}.
However, little is understood about such improvement at the instance (datapoint) level.
Are larger models uniformly better? In other words, are larger pretrained models better at every instance, or are they better at some instances, but worse at others?

Prior works hint at differing answers.
\citet{hendrycks-etal-2020-pretrained} and \citet{desai-durrett-2020-calibration} find that larger pretrained models consistently improve out-of-distribution performance, which implies that they might be uniformly better at a finer level. 
\citet{henighan2020scaling} claim that larger pretrained image models have lower downstream classification loss for the majority of instances, and they predict this trend to be true for other data modalities (e.g. text).
On the other hand, \citet{pmlr-v119-sagawa20a} find that larger non-pretrained models perform worse on rare subgroups;
if this result generalizes to pretrained language models, larger models will not be uniformly better.
Despite all the indirect evidence, it is still inconclusive how many instances larger pretrained models perform worse on.

A naïve solution is to finetune a larger model, compare it to a smaller one, and find instances where the larger model is worse.
However, this approach is flawed, since model predictions are \textbf{noisy at the instance level}.
On MNLI in-domain development set, even the same architecture with different finetuning seeds leads to different predictions on $\sim$8\% of the instances.
This is due to under-specification \cite{d2020underspecification}, where there are multiple different solutions that can minimize the training loss. 
Since the accuracy improvement from our BERT-$\basesize$\footnote{This is not the original release by \citet{devlin-etal-2019-bert}; we pretrained models ourselves.}
to BERT-$\largesize$ is 2\%, most signals across different model sizes will be dominated by noise due to random seeds.

\begin{figure*}[t]
    \centering
    \includegraphics[width=\textwidth]{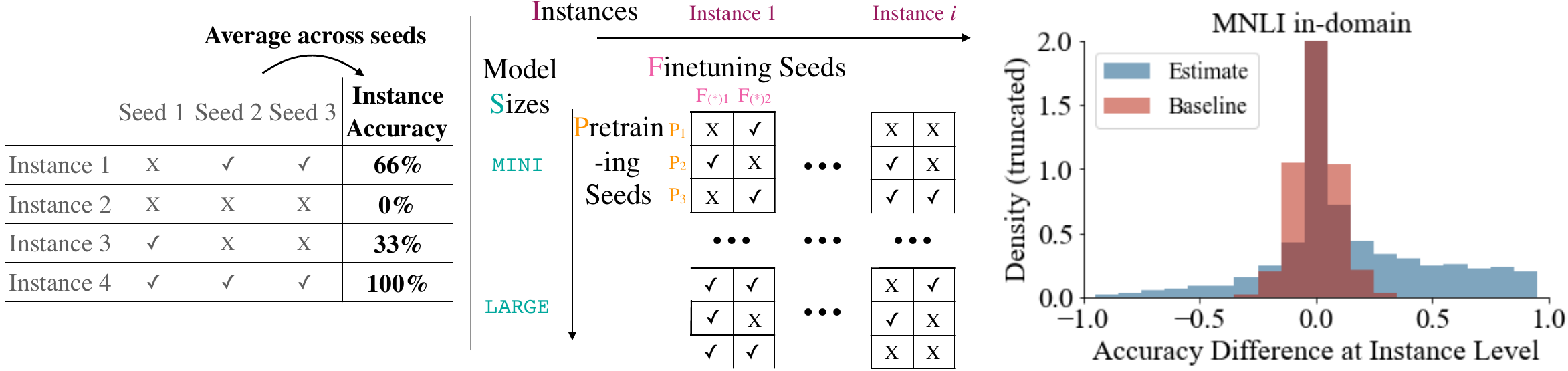}
    \caption{\textbf{Left}: Each column represents the same architecture trained with a different seed.
    We calculate accuracy for each instance (row) by averaging across seeds (column), while it is usually calculated for each model by averaging across instances.
    \textbf{Middle}: A visual layout of the model predictions we obtain, which is a binary-valued tensor with 4 axes: model size $s$, instance $i$, pretraining seeds $P$ and finetuning seeds $F$.
    \textbf{Right}: for each instance, we calculate the accuracy gain from $\minisize$ to $\largesize$ and plot the histogram in blue, along with a random baseline in red. Since the blue distribution has a bigger left tail, smaller models are better at some instances.
    }
    \label{fig:main-fig}
\end{figure*}

To account for the noise in pretraining and finetuning, we define \textit{instance accuracy} as ``how often a model correctly predicts an instance" (Figure \ref{fig:main-fig} left) in expectation across pretraining and finetuning seeds. 
We estimate this quantity by pretraining 10 models with different seeds, finetuning 5 times for each pretrained models (Figure \ref{fig:main-fig} middle), and averaging across them.

However, this estimate is still inexact, and we might falsely observe smaller models to be better at some instances by chance.
Hence, we propose a random baseline to estimate the fraction of \textit{false discoveries} (Section \ref{sec:instance-acc-cmp}, Figure \ref{fig:main-fig} right) and formally upper-bound the false discoveries in Section \ref{sec:formal-guarantee}. 
Our method provides a better upper bound than the classical Benjamini-Hochberg procedure with Fisher's exact test.

Using the 50 models for each size and our improved statistical tool, we find that, on the MNLI in-domain development set, the accuracy ``decays" from BERT-\largesize to BERT-\minisize on at least $\sim$4\% of the instances, which is significant given that the improvement in overall accuracy is 10\%. 
These decaying instances contain more controversial or wrong labels, but also correct ones (Section \ref{sec:understanddecay}).
Therefore, larger pretrained language models are not uniformly better.

We make other interesting discoveries at the instance level.
Section \ref{sec:predictability} finds that instance-level accuracy has momentum: improvement from \minisize to \mediumsize correlates with improvement from \mediumsize to \largesize.
Additionally, Section \ref{sec:main-variance} attributes variance of model predictions to pretraining and finetuning random seeds, and finds that finetuning seeds cause more variance for larger models.
Our findings suggest that instance-level predictions provide a rich source of information; we therefore recommend that researchers supplement model weights with model predictions.
In this spirit, we release all the pretrained models, model predictions, and code here: \url{https://github.com/ruiqi-zhong/acl2021-instance-level}.

\section{Data, Models, and Predictions}
\label{sec:prediction-data}

To investigate model behavior, we considered different sizes of the BERT architecture and finetuned them on Quora Question Pairs (QQP\footnote{https://www.quora.com/q/quoradata/First-Quora-Dataset-Release-Question-Pairs}), Multi-Genre Natural Language Inference (MNLI; \citet{williams-etal-2020-predicting}), and the Stanford Sentiment Treebank (SST-2; \citet{socher-etal-2013-recursive}). 
To account for pretraining and finetuning noise, we averaged over multiple random initializations and training data order,  and thus needed to pretrain our own models rather than downloading off the internet. 
Following \citet{turc2019} we trained 5 architectures of increasing size: \minisize (L4/H256, 4 Layers with hidden dimension 256), \smallsize (L4/H512), \mediumsize (L8/H512), \basesize  (L12/H768), and \largesize (L24/H1024). 
For each architecture we pre-trained models with 10 different random seeds and fine-tuned each of them 5 times (50 total) on each task; see Figure \ref{fig:main-fig} middle.
Since pretraining is computationally expensive, we reduced the context size during pretraining from 512 to 128 and compensated by increasing training steps from 1M to 2M. 
Appendix \ref{sec:app-prediction-data} includes more details about pretraining and finetuning and their computational cost, and Appendix \ref{sec:compare-to-orig} verifies that our cost-saving changes do not affect accuracy qualitatively. 


\paragraph{Notation.}
We use $i$ to index an instance in the evaluation set, $s$ for model sizes, $P$ for pretraining seeds and $F$ for finetuning seeds. 
$c$ is a random variable of value 0 or 1 to indicate whether the prediction is correct.
Given the pretraining seed $P$ and the finetuning seed $F$, $c^{s}_{i}= 1$  if the model of size $s$ is correct on instance $i$, 0 otherwise. 
To keep the notation uncluttered, we sometimes omit these superscripts or subscripts if they can be inferred from context.

Unless otherwise noted, we present results on the MNLI in-domain development set in the main paper.

\section{Comparing Instance Accuracy} \label{sec:instance-acc-cmp}

To find the instances where larger models are worse, a naïve approach is to finetune a larger pretrained model, compare it to a smaller one, and find instances where the larger is incorrect but the smaller is correct.
Under this approach, BERT-\largesize is worse than BERT-\basesize on 4.5\% of the instances and better on 7\%, giving an overall accuracy improvement of 2.5\%. 

However, this result is misleading: even if we compare two BERT-\basesize model with different finetuning seeds, their predictions differ on 8\% of the instances, while their accuracies differ only by 0.1\%; Table~\ref{tab:diff} reports this baseline randomness across model sizes. Changing the pretraining seed also changes around $2\%$ additional predictions beyond finetuning.

Table \ref{tab:diff} also reports the standard deviation of overall accuracy, which is about 40 times smaller. 
Such stability starkly contrasts with the noisiness at the instance level, which poses a unique challenge.

\begin{table}[]
    \centering
    \begin{tabular}{lccc}
    \hline
        {} &  $\difff$ &  $\diffp$ &  $\avgstd$ \\
        \hline
        \minisize   &  7.2\% & 10.7\% &  0.2\% \\
        \smallsize  &   7.2\% & 10.7\% & 0.3\% \\
        \mediumsize &   8.0\% & 10.7\% &  0.3\% \\
        \basesize   &   8.5\% & 10.6\% & 0.2\% \\
        \largesize  &   8.6\% & 10.1\% &  0.2\% \\
        \hline
    \end{tabular}
    \caption{Larger model sizes are at the bottom rows. 
    $\difff$: how much do the predictions differ, if two models have the same pretraining seed but different finetuning seeds $F$? $\diffp$: the difference if the pretraining seeds $P$ are different.
    $\avgstd$: the standard deviation of overall accuracy, around 40 times smaller than $\difff$.
 }
    \label{tab:diff}
\end{table}

\paragraph{Instance-Level Metrics}
To reflect this noisiness, we define the \textit{instance accuracy} $\acc^{s}_{i}$ to be 
how often models of size $s$ predict instance $i$ correctly,
\begin{equation} \label{eq:instance-acc-def}
    \acc_{i}^{s} := \mathbb{E}_{P,F}[c_{i}^{s}].
\end{equation}
The expectation is taken with respect to the pretraining and finetuning randomness $P$ and $F$. 
We estimate $\acc_{i}^s$ via the empirical average $\hat{\acc}_{i}^{s}$ accross 10 pretraining $\times$ 5 finetuning runs. 

\begin{figure*}[t]
    \subfloat[\centering \basesize vs. \largesize, $\acc$]{{\includegraphics[width=0.60\columnwidth]{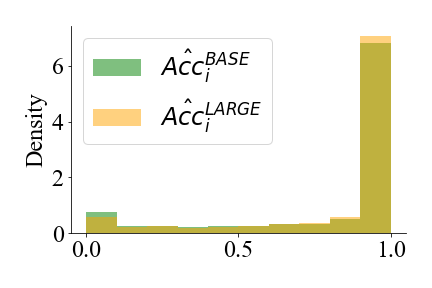}}}
    \space
    \subfloat[\centering \basesize vs. \largesize, $\Delta\acc$]{{\includegraphics[width=0.65\columnwidth]{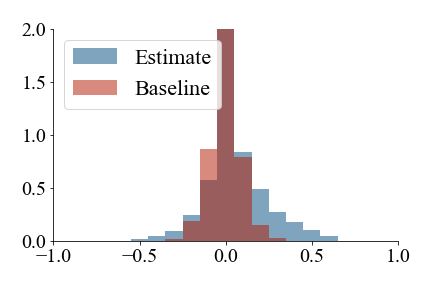} }}%
    \space
    \subfloat[\centering \minisize vs. \largesize, $\Delta\acc$]{{\includegraphics[width=0.65\columnwidth]{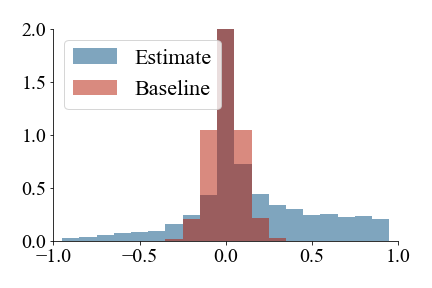} }}%
    \caption{
    (a) The distribution of instance accuracy $\hat{\acc}_{i}$.
    (b, c) Histogram of instance difference estimate (x-axis), $\hat{\Delta\acc}$ (blue) and its baseline $\Delta\acc'$ (red) compares \basesize and \largesize.
    To better visualize, we truncated the density (y-axis) above 2. 
    Since the blue histogram has a larger left tail than the red one, there are indeed instances where larger models are worse.
    }%
    \centering
    \label{fig:main-instance-metric}
\end{figure*}

We histogram $\hat{\acc}_i^s$ in Figure \ref{fig:main-instance-metric} (a). 
On most instances the model always predicts correctly or incorrectly ($\hat{\acc} = 0$ or $1$), but a sizable fraction of accuracies lie between the two extremes.

Recall that our goal is to find instances where larger models are less accurate, which we refer to as \textit{decaying} instances. 
We therefore study the \textit{instance difference} between two model sizes $s_1$ and 
$s_2$, 
defined as 
\begin{equation} \label{eq:instance-diff-def}
    ^{s_{1}}_{s_{2}}\Delta\acc_{i}  
    := \acc^{s_{2}}_{i}- \acc^{s_{1}}_{i},
\end{equation}
which is estimated by the difference between the accuracy estimates $\hat{\acc}_{i}^{s}$, i.e.

\begin{equation} \label{instance-difference-estimator-def}
    ^{s_{1}}_{s_{2}}\hat{\Delta\acc_{i}}  
    := \hat{\acc}^{s_{2}}_{i}- \hat{\acc}^{s_{1}}_{i}.
\end{equation}

We histogram $^{\basesize}_{\largesize}\hat{\Delta\acc_{i}}$ in Figure \ref{fig:main-instance-metric} (b). 
We observe a unimodal distribution centered near 0, with tails on both sides.
Therefore, the estimated differences for some instances are negative.

\begin{figure}[h]
    \centering
    \includegraphics[width=\columnwidth]{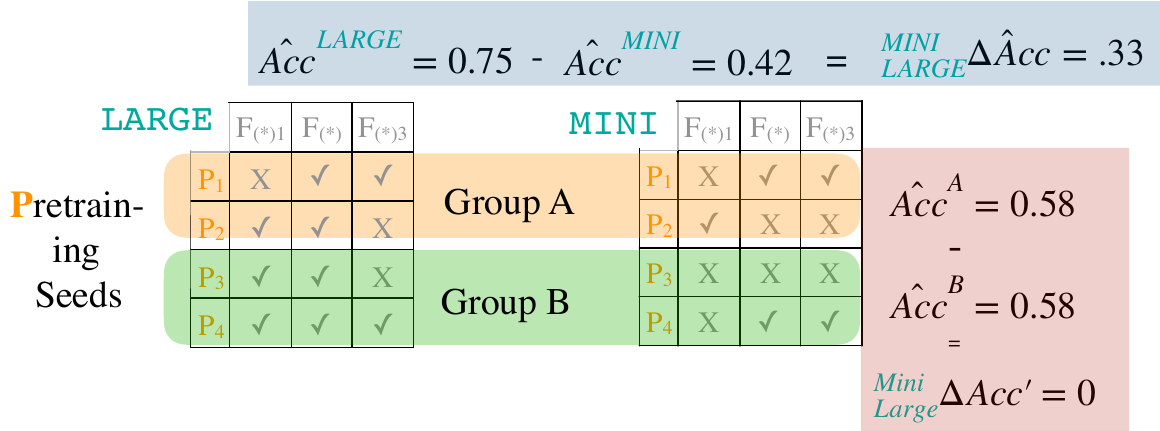}
    \caption{The tables are model predictions with visual notations established in Figure \ref{fig:main-fig} middle. $\hat{\Delta\acc}$ (blue) is the mean difference between the left and the right table, each corresponding to a model size. The random baseline $\Delta\acc'$ (red) is the mean difference between group $A$ (orange) cells and group $B$ (green), which are identically and independently distributed.}
    \label{fig:mixingestimator}
\end{figure}

However, due to estimation noise, we might falsely observe this accuracy decay by chance.
Therefore, we introduce a random baseline $^{s_{1}}_{s_{2}}\Delta\acc'$ to control for these false discoveries. 
Recall that we have 10 smaller pretrained models and 10 larger ones.
Our baseline splits these into a group $A$ of $5$ smaller + $5$ larger, and another group $B$ of the remaining $5 + 5$. Then the empirical accuracies $\hat{\acc}^A$ and $\hat{\acc}^B$ are identically distributed, so we take our baseline 
$^{s_{1}}_{s_{2}}\Delta\acc'$ to be the difference  $\hat{\acc}^{A}-\hat{\acc}^{B}$.
We visualize and compare how to calculate $^{s_{1}}_{s_{2}}\hat{\Delta\acc}$ and $^{s_{1}}_{s_{2}}\Delta\acc'$ in Figure \ref{fig:mixingestimator}.

We histogram this baseline $^{\basesize}_{\largesize}\Delta\acc'$ in Figure \ref{fig:main-instance-metric} (b), and find that our noisy estimate $^{\basesize}_{\largesize}\hat{\Delta\acc}$ has a larger left tail than the baseline.
This suggests that decaying instances exist.
We similarly compare \minisize to \largesize in Figure \ref{fig:main-instance-metric} (c) and find an even larger left tail. 

\section{Quantifying the Decaying Instances} \label{sec:formal-guarantee}
The left tail of $\hat{\Delta\acc}$ noisily estimates the fraction of decaying instances, and the left tail of the random baseline $\Delta\acc'$ counts the false discovery fraction due to the noise.
Intuitively, the true fraction of decaying instances can be captured by the difference of these left tails, and we formally quantify this below.

Suppose instance $i$ is drawn from the empirical evaluation distribution. Then we can define the true decaying fraction $\decay$ as 

\begin{equation} \label{eq:true-decay}
    \decay := \mathbb{P}_{i}[\Delta \acc_{i} < 0].
\end{equation}

Since $\Delta\acc_{i}$ is not directly observable and $\hat{\Delta\acc}_{i}$ is noisy, we add a buffer and only consider instances with $\hat{\Delta\acc}_{i}\leq t$, which makes it more likely (but still uncertain) that the true $\Delta\acc_{i} < 0$.
We denote this ``discovery fraction" $\hat{\decay}(t)$ as

\begin{equation} \label{eq:discovery-decay}
    \hat{\decay (t}) := \mathbb{P}_{i}[\hat{\Delta\acc}_{i} \leq t].
\end{equation}

Similarly, we define a baseline control (false discovery fraction) $\decay' (t)  := \mathbb{P}_{i}[\Delta\acc'_{i} \leq t]$.
Hence, $\hat{\decay}$ and $\decay'$ are the cumulative distribution function of $\hat{\Delta\acc}$ and $\Delta\acc'$ (Figure \ref{fig:cum_instance_diff}).

We have the following theorem, which we formally state and prove in Appendix \ref{sec:our-fdr-estimate}:

\begin{theorem} \label{thm:decaylowerbound}
(Informal) If all the random seeds are independent, then for all thresholds $t$,
\begin{equation}
    \decay \geq \mathbb{E}[\hat{\decay}(t) - \decay'(t)]
\end{equation}
\end{theorem}

\paragraph{Proof Sketch} 

Suppose we observe $c^{s_{1}}_{R_{1\dots 2k}}$ and $c^{s_{2}}_{R_{2k+1\dots 4k}}$, where there are $2k$ different random seeds for each model size \footnote{We assumed even number of random seeds since we will mix half of the models from each size to compute the random baseline}. 
Then
\begin{equation} 
    \hat{\Delta \acc}_{i} := \frac{1}{2k}(\sum^{2k}_{j=1}c^{s_{1}}_{R_{j},i} - \sum^{4k}_{j=2k+1}c^{s_{2}}_{R_{j},i}),
\end{equation}

and hence the discovery rate $\hat{\decay}(t)$ is defined as 
\begin{align}
    \hat{\decay}(t) := \frac{1}{|\mathcal{T}|}\sum_{i=1}^{|\mathcal{T}|}\mathbf{1}[\hat{\Delta \acc} \leq t].
\end{align}

For the random baseline estimator, we have
\begin{align}
    \Delta \acc'_{i} &:= \frac{1}{2k}(\sum^{k}_{j=1}c^{s_{1}}_{R_{j}, i} + \sum^{3k}_{j=2k+1}c^{s_{2}}_{R_{j}, i} \\
    &- \sum^{2k}_{j=k+1}c^{s_{1}}_{R_{j},i} - \sum^{4k}_{j=3k+1}c^{s_{2}}_{R_{j},i} \nonumber
    ),
\end{align}

and the false discovery control $\decay'$ is defined as 
\begin{align}
    \decay'(t) := \frac{1}{|\mathcal{T}|}\sum_{i=1}^{|\mathcal{T}|}\mathbf{1}[\Delta \acc'_{i} \leq t].
\end{align}

Formally, the theorem states that
\begin{equation}
    \decay \geq \mathbb{E}_{R_{1}\dots R_{4k}}[\hat{\decay}(t) - \decay'(t)],
\end{equation}
which is equivalent to
\begin{align}
    \sum_{i=1}^{|\mathcal{T}|}(\mathbf{1}[\Delta \acc_{i} < 0] - \mathbb{P}[\hat{\Delta \acc}_{i} \leq t] \\+ \mathbb{P}[\Delta \acc'_{i} \leq t]) \geq 0 \nonumber
\end{align}

Hence, we can declare victory if we can prove that for all $i$, if $\Delta \acc_{i} \geq 0$, 
\begin{align}
\mathbb{P}[\Delta \acc'_{i} \leq t] \geq \mathbb{P}[\hat{\Delta \acc}_{i} \leq t]. \nonumber
\end{align}
This is easy to see, since $\Delta \acc'_{i}$ and $\hat{\Delta \acc}_{i}$ are both binomial distributions with the same $n$, but the first has a larger rate. 
\footnote{More details are in Appendix \ref{sec:our-fdr-estimate}.}
$\square$

Roughly speaking, the true decaying fraction is at least the difference between $\hat{\decay}(t)$ and $\decay'(t)$ at every threshold $t$.
Therefore, we take the maximum difference between $\hat{\decay}(t)$ and $\decay'(t)$ to lower-bound the fraction of decaying instances.\footnote{Adaptively picking the best threshold $t$ depending on the data may incur a slight upward bias.
Appendix \ref{sec:adaptive-thresholds} estimates that the relative bias is at most 10\% using a bootstrap method.}
For example, Figure \ref{fig:cum_instance_diff} estimates the true decaying fraction between \minisize and \largesize to be at least 6\%.

\begin{figure}[tb]
    \centering
    \includegraphics[width=\columnwidth]{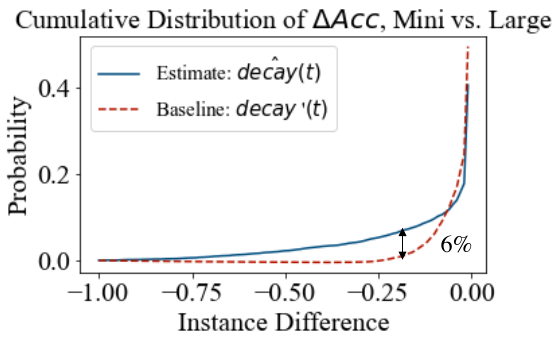}
    \caption{The cumulative distribution function of the histogram in Figure \ref{fig:main-instance-metric} (c); only the negative x-axis is shown because it corresponds to decays.
    The maximum difference between the two curves (6\%) is a lower bound of the true decaying fraction.
    }
    \label{fig:cum_instance_diff}
\end{figure}

We compute this lower bound for other pairs of model sizes in Table \ref{tab:main-cmp}, and the full results across other tasks and model size pairs are in Appendix \ref{app:more-instance-cmp}.
In all of these settings we find a non-zero fraction of decaying instances, and larger model size differences usually lead to more decaying instances.

Unfortunately, 
applying Theorem \ref{thm:decaylowerbound} as above is not fully rigorous, 
since some finetuning runs share the same pretraining seeds and hence are dependent.\footnote{Although we anticipate such dependencies do not cause a substantial difference, as discussed in Appendix \ref{sec:independent-seed-assumption}.}
To obtain a statistically rigorous lower bound, we slightly modify our target of interest. 
Instead of examining individual finetuning runs, we ensemble our model across 5 different finetuning runs for each pretraining seed;
these predictions are essentially the same as individual finetuning runs, except that the finetuning randomness is averaged out.
Hence we obtain 10 independent sets of model predictions with different random seeds, which allows us to apply Theorem \ref{thm:decaylowerbound}.

We compare \minisize to \largesize using these ensembles and report the discovery $\hat{\decay}$ and the baseline $\decay'$ in Table \ref{tab:minivs.large}. 
Taking the maximum difference across thresholds, we estimate at least $\sim$4\% of decaying instances. 
This estimate is lower than the previous 6\% estimate, which used the full set of 50 models' predictions assuming they were independent.
However, this is still a meaningful amount, given that the overall accuracy improvement from \minisize to \largesize is 10\%.

\begin{table}[t]
    \centering
        \begin{tabular}{|l|c|c|c|c|c|}
        \hline
        $s_{1}\setminus s_{2}$ &  \minisize & \smallsize &  \basesize &  \largesize \\
        \hline
        \minisize  &   N/A &  9\% &    18\% &  21\% \\
        \hline
        \smallsize   & 3\% &    N/A &    14\% &  18\% \\
        \hline
        \basesize   & 6\% &  5\% &      N/A &  10\% \\
        \hline
        \largesize  & 6\% &  5\% &    \phantom{0}2\% &    N/A \\
        \hline
    \end{tabular}
    \caption{We lower-bound the fraction of instances that improve when model size changes from $s_{1}$ (row) to $s_{2}$ (column). For example, when model size decreases from \largesize to \minisize, 6\% of instances improve (i.e. decays).}
    \label{tab:main-cmp}
\end{table}

\begin{table}[t]
    \centering
    \begin{tabular}{cccc}
    \hline
    Threshold & $\hat{\decay}$ & $\decay'$ & Diff \\
    \hline
    $t = -0.4$ & 4.22\% & \num{3.49e-3} & 3.87\% \\
    \dots & \dots & \dots & \dots \\
    $t = -0.9$ & 0.91\% & \num{1.44e-7} & 0.91\% \\
    $t = -1.0$ & 0.48\% & \num{2.06e-8} & 0.48\% \\
    \hline
    \end{tabular}
    \caption{Comparing \minisize vs. \largesize by calculating the discovery fraction $\hat{\decay}$, the false discovery control $\decay'$, and their difference (Diff) under different thresholds $t$.
    \largesize is worse on at least $\sim$4\% (maximum Diff) of instances.
    }
    \label{tab:minivs.large}
\end{table}

\subsection{Fisher's Test + Benjamini-Hochberg}
Here is a more classical approach to lower-bound the decaying fraction. 
For each instance, we compute a significance level $\alpha$ under the null hypothesis that the larger model is better, using Fisher's exact test. We sort the significance levels ascendingly, and call the $p^{th}$ percentile $\alpha_{p}$. 
Then we pick a false discovery rate $q$ (say, 25\%), find the largest $p$ s.t. $\alpha_{p} < pq$, and estimate the decaying fraction to be at least $p(1 - q)$. 
This calculation is known as the Benjamini-Hochberg procedure \cite{benjamini1995controlling}.

To compare our method with this classical approach, we estimate the lower bound of the decaying fraction for different pairs of model sizes with different numbers of pretrained models available.
To make sure our choice of the false discovery rate $q$ does not bias against the classical approach, we adaptively choose $q$ to maximize its performance.
Appendix~\ref{sec:cmp-signifiance} includes the full results and Table \ref{tab:main-compare-ours-bh} is a representative subset.

We find that our approach is more powerful, particularly when the true decaying fraction is likely to be small and only a few models are available, which is usually the regime of interest. 
For example, across all pairs of model sizes, our approach only needs 2 random seeds (i.e. pretrained models) to provide a non-zero lower bound on the decaying fraction, while the classical approach sometimes fails to do this even with 10 seeds. 
Intuitively, when fewer seeds are available, the smallest possible significance level for each instance is larger than the decaying fraction, hence hurting the classical approach. 

\begin{table}[t]
    \centering
    \begin{tabular}{lllccccc}
    \hline
    s1 &      s2 &  &     2 &      6 &    10 \\
    \hline
    \minisize &   \largesize &   ours & 1.9\% &  3.1\% &  4.0\% \\
    \minisize &   \largesize &     BH & 0.0\% &  0.9\%  & 1.9\% \\
    \hline
    \basesize &   \largesize &   ours & 0.4\% &  0.9\% &  1.2\% \\
    \basesize &   \largesize &     BH & 0.0\% & 0.0\% & 0.0\% \\
    \hline
    \end{tabular}
    \caption{
    We compare our method to the Fisher's exact test + Benjamin-Hochberg (BH) procedure described in Section \ref{sec:formal-guarantee}.
    For all different model size pairs and number of pretrained models available, ours always provides a higher (better) lower bound of the decaying fraction. 
    }
    \label{tab:main-compare-ours-bh}
\end{table}
\subsection{Understanding the Decaying Instances} \label{sec:understanddecay}
We next manually examine the decaying instances to see whether we can find any interpretable patterns.
One hypothesis is that all the decaying fractions are in fact mislabeled, and hence larger models are not in fact worse on any instances.

To investigate this hypothesis, we examined the group of instances where  $^{\minisize}_{\largesize}\hat{\Delta\acc}_{i} \leq -0.9$.
\minisize is almost always correct on these instances, while \largesize is almost always wrong, and the false discovery fraction is tiny. 
For each instance, 
we manually categorize it as either: 1) Correct, if the label is correct, 2) Fine, if the label might be controversial but we could see a reason why this label is reasonable, 3) Wrong, if the label is wrong, or 4) Unsure, if we are unsure about how to label this instance.
Each time we annotate, with 50\% probability we randomly sample either a decaying instance or an instance from the remaining dataset as a control. We are blind to which group it comes from.

For each task of MNLI, QQP, and SST-2,  the first author annotated 100 instances (decay + control group) (Table~\ref{tab:manual}).
We present all the annotated decaying instances in Appendix \ref{sec:decaying-instances-all}.

\paragraph{Conclusion} 
We find that the decaying fraction has more wrong or controversial labels, compared to the remaining instances. 
However, even after we adjust for the fraction of incorrect labels, the $\decay$ fraction 
still exceeds the false discovery control. This implies that \minisize models are better than \largesize models on some correctly labeled instances.
The second author followed the same procedure and reproduced the same qualitative results.

However, we cannot find an interpretable pattern for these correctly labeled decaying instances by simply eyeballing. 
We discuss future directions to discover interpretable categories in Section \ref{sec:conclusion}.

\begin{table}[]
    \centering
    \begin{tabular}{rcccc}
     & Correct & Fine & Wrong & Unsure  \\
     \hline
   MNLI$^{D}$ & 66\% & 17\% & \phantom{0}9\% & \phantom{0}5\% \\
   MNLI$^{C}$ & 86\% & \phantom{0}5\% & 5\% & \phantom{0}1\% \\
   \hline
   SST-2$^{D}$ & 55\% & 8\% & 10\% & 25\%\\
   SST-2$^{C}$ & 88\% & \phantom{0}4\% & \phantom{0}0\% & \phantom{0}6\% \\
   \hline
   QQP$^{D}$  & 60\% & 26\% & 10\% & \phantom{0}2\% \\
   QQP$^{C}$  & 87\% & 10\% & \phantom{0}1\% & \phantom{0}0\% \\
   \hline
\end{tabular}
    \caption{\minisize vs. \largesize.  
    We examine whether there are mislabels for the \textbf{D}ecaying fractions  (superscript $^D$) and the rest of the dataset (\textbf{C}ontrol group $^C$). 
    The decaying fraction contains more mislabels, but includes correct labels as well.
    }
    \label{tab:manual}
\end{table}
\section{Correlation of Instance Difference} \label{sec:predictability}
\begin{table*}[t]
    \centering
    \begin{tabular}{c|ccccccccccc}
     $(s_{1}, s_{2}, s_{3})\downarrow$ Buckets$\rightarrow$ & 0.10 & 0.20 & 0.30 & 0.40 & 0.50 & 0.60 & 0.70 & 0.80 & 0.90 & 1.00 \\
    \hline
    $\smallsize,\mediumsize,\basesize$ & 0.07 &  0.22 &  0.29 &  0.40 &  0.35 &  0.33 &  0.38 &  0.27 &  0.24 &  0.13 \\
    $\minisize,\mediumsize,\largesize$ & 0.03 &  0.15 &  0.18 &  0.33 &  0.17 &  0.16 &  0.22 &  0.20 &  0.19 &  0.09 \\
    \hline
    \end{tabular}
    \caption{
    Each row corresponds to a triplet of model sizes.
    Each column $t$ represents a bucket that contains instances with $\hat{\acc}^{s_{2}} \in [t-0.1,t]$. 
    Within each bucket, we calculate the Pearson correlation coefficient between the estimated accuracy improvements: $^{s_{1}}_{s_{2}}\hat{\Delta \acc}$ and $^{s_{2}}_{s_{3}}\hat{\Delta \acc}$.
    These correlations are positive and become higher when model size differences are small. 
    }
    \label{tab:main-momentum}
\end{table*}

We next investigate whether there is a momentum of instance accuracy increase:
for example, if the instance accuracy improves from $\minisize (s_{1})$ to $\mediumsize(s_{2})$, is it more likely to improve from $\mediumsize (s_{2})$ to $\largesize (s_{3})$?

The na\"{i}ve approach is to calculate the Pearson correlation coefficient between $^{\minisize}_{\mediumsize}\hat{\Delta\acc}$ and $^{\mediumsize}_{\largesize}\hat{\Delta\acc}$, and we find the correlation to be zero. 
However, this is partly an artifact of accuracies being bounded in $[0,1]$. 
If $\mediumsize$ drastically improves over $\minisize$ from 0 to 1,
there is no room for $\largesize$ to improve over $\mediumsize$. 
To remove this inherent negative correlation, we calculate the correlation conditioned on the accuracy of the middle-sized model, $\hat{\acc}^{\mediumsize}$.

Therefore, we bucket instances by their estimated \mediumsize accuracy into intervals of size 0.1, and we find the correlation to be positive within each bucket (Table \ref{tab:main-momentum}, row 2).
This fixes the problem with the naïve approach by getting rid of the negative correlation, which could have misled us to believe that improvements by larger models are uncorrelated.

We additionally find that the correlations between improvements become stronger when model size differences are smaller.
Table \ref{tab:main-momentum} row 1 reports results for another model size triplet with smaller size difference, i.e. ($s_{1}$, $s_{2}$, $s_{3}$) = ($\smallsize$, $\mediumsize$, $\basesize$), and the correlation is larger for all buckets. 
Results for more tasks and size triplets are in Appendix \ref{sec:more-on-momentum} and the same conclusions hold qualitatively. 


\section{Variance at the Instance Level} \label{sec:main-variance}
Section \ref{sec:instance-acc-cmp} found that the overall accuracy has relatively low variance, but model predictions are noisy.
This section formally analyzes variance at the instance level.
For each instance, we decompose its loss into three components: Bias$^{2}$, variance due to pretraining randomness, and variance due to finetuning randomness. 
Formally, we consider the $0/1$ loss:
\begin{equation}
    \mathcal{L}_{i} := 1 - c_{i} = (1 - c_{i})^{2},
\end{equation}
where $c_{i}$ is a random variable 0/1 indicating whether the prediction is correct or incorrect, with respect to randomness in pretraining and finetuning.
Therefore, by bias-variance decomposition and total variance decomposition, we have
\begin{align}
    \mathcal{L}_{i} = \biassquared_{i} + \pretrainvar_{i} + \finetunevar_{i},
\end{align}
where, by using $P$ and $F$ as pretraining and finetuning random seeds:
\begin{align}
    \biassquared_{i} &:= (1 - \mathbb{E}_{P,F}[c_{i}])^{2}, \\
    \pretrainvar_{i} &:= \Var_{P}[\mathbb{E}_{F}[c_{i}]], \nonumber\\
     \finetunevar_{i} &:= \mathbb{E}_{P}[\Var_{F}[c_{i}]], \nonumber
\end{align}
capturing ``how wrong is the average prediction", variance due to pretraining, and variance due to finetuning seeds, respectively. 

We can directly estimate $\finetunevar$ by first calculating the sample variance across finetuning runs for each pretraining seed, and then averaging the variances across the pretraining seeds.
Estimating $\pretrainvar$ is more complicated. 
A na\"{i}ve approach is to 
calculate the empirical variance, across pretraining seeds, of the average accuracy across finetuning seeds.
However, the estimated average accuracy for each pretraining seed is noisy itself, which causes an upward bias on the $\pretrainvar$ estimate.
We correct this bias by estimating the variance of the estimated average accuracy and subtracting it from the na\"{i}ve estimate; see 
Appendix~\ref{sec:decompose} for details, as well as a generalization to more than two sources of randomness.
Finally, we estimate $\biassquared$ by subtracting the two variance estimates from the estimated loss.

\begin{table}[]
    \centering
    \begin{tabular}{lcccc}
\hline
 &  $\biassquared$ & $\pretrainvar$ &  $\finetunevar$ \\
\hline
    \minisize   &   0.203 & 0.017 & 0.036 \\
    \smallsize  &   0.179 & 0.017 & 0.036 \\
    \mediumsize &   0.157 & 0.014 & 0.040 \\
    \basesize   &   0.134 & 0.010 & 0.043 \\
    \largesize  &   0.111 & 0.007 & 0.043 \\
\hline
\end{tabular}
    \caption{The bias, pretraining variance, and finetuning variance for each model size, averaged across all test instances. Finetuning variance is much larger than pretraining variance; larger models have larger finetuning variance.
    }
    \label{tab:varianes-main}
\end{table}

\begin{figure}[tbh]
    \centering
    \includegraphics[width=\columnwidth]{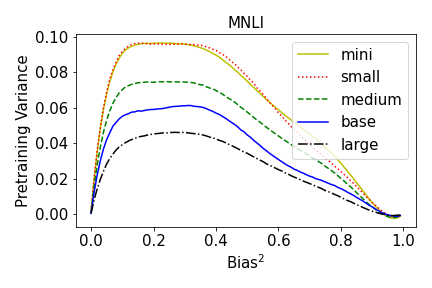}
    \caption{The pretraining variance conditioned on $\biassquared$ (the level of correctness). Each curve represents a model size. Larger models have lower pretraining variance across all levels of bias.}
    \label{fig:main-bias-var-curve}
\end{figure}

For each of these three quantities, $\biassquared$, $\pretrainvar$ and $\finetunevar$, we estimate it for each instance, average it across all instances in the evaluation set, and report it in Table \ref{tab:varianes-main}. 
The variances at the instance level are much larger than the variance of overall accuracy, by a factor of 1000.

We may conclude from Table \ref{tab:varianes-main} that larger models have larger finetuning variance and smaller pretraining variance.
However, lower bias also inherently implies lower variance.
To see this, suppose a model has perfect accuracy and hence zero bias; then it always predicts the same label (the correct one) and hence has zero variance. 
This might favor larger models and ``underestimate" their variance, since they have lower bias.
Therefore, we calculate and compare the variances conditioned on the bias, i.e. $\pretrainvar(b^{2}) := \mathbb{E}_{i}[\pretrainvar_{i}|\biassquared_{i}=b^2]$.

We estimate $\pretrainvar^{s}(b^{2})$ using Gaussian process regression and plot it against $b^{2}$ in Figure \ref{fig:main-bias-var-curve}. We find that larger models still have lower pretraining variance across all levels of bias on the specific task of MNLI under the 0/1 loss. 
To further check whether our conclusions are general, we tested them on other tasks and under the squared loss $\mathcal{L}_{i}:=(1-p_{i})^{2}$, where $p_{i}$ is the probability assigned to the correct class. 
Below are the conclusions that generally hold across different tasks and loss functions.

\paragraph{Conclusion} We find that 1) larger models have larger finetuning variance, 2) \largesize has smaller pretraining variance than \basesize; however, the ordering between other sizes varies across tasks and losses, and 3) finetuning variance is 2$-$8 times as large as pretraining variance, and the ratio is bigger for larger models.

\section{Discussion and Future Directions} \label{sec:conclusion}
To investigate model behaviors at the instance level, we produced massive amounts of model predictions in Section \ref{sec:prediction-data} and treated them as raw data.
To extract insights from them, we developed better metrics and statistical tools, including a new method to control the false discoveries, an unbiased estimator for the decomposed variances, and metrics that compute variance and correlation of improvements conditioned on instance accuracy. 
We find that larger pretrained models are indeed worse on a non-trivial fraction of instances and have higher variance due to finetuning seeds;
additionally, instance accuracy improvements from \minisize to \mediumsize correlate with improvements from \mediumsize to \largesize.

Overall, we treated model prediction data as the central object and built analysis tools around them to obtain a finer understanding of model performance.
We therefore refer to this paradigm as ``\textbf{instance level understanding as data mining}". 
We discuss three key factors for this paradigm to thrive:
1) scalability and the cost of obtaining prediction data, 2) other information to collect for each instance, and 3) better statistical tools. 
We analyze each of these aspects below.

\paragraph{Scalability and Cost of Data}
Data mining is more powerful with more data. How easy is it to obtain more model predictions?
In our paper, the main bottleneck is pretraining. 
However, once the pretrained models are released, individual researchers can download them and only need to repeat the cheaper finetuning procedure.

Furthermore, model prediction data are under-shared:
while many recent research papers share their code or even model weights to help reproduce the results, it is not yet a standard practice to share all the model predictions.
Since many researches follow almost the same recipe of pretraining and finetuning \cite{mccoy-etal-2020-berts, desai-durrett-2020-calibration, dodge2020fine}, much computation can be saved if model predictions are shared.
On the other hand, as the state of the art model size is increasing at a staggering speed\footnote{e.g. BERT \cite{devlin-etal-2019-bert} has 340M parameters, while Switch-Transformer has over 1 trillion parameters \cite{fedus2021switch}. }, 
most researchers will not be able to run inference on a single instance.
The trend that models are becoming larger and more similar necessitate more prediction sharing.

\paragraph{Meta-Labels and Other Predictions} 
Data mining is more powerful with more types of information.
One way to add information to each instance is to assign ``meta-labels". 
In the HANS \cite{mccoy-etal-2019-right} dataset, the authors tag each instance with a heuristic \footnote{For example, ``the label [entailment] is likely if the premise and the hypothesis have significant lexical overlap".} that holds for the training distribution but fails on this instance.
\citet{naik18coling} and \citet{ribeiro-etal-2020-beyond} associate each instance with a particular stress test type or subgroup, for example, whether the instance requires the model to reason numerically or handle negations.
\citet{nie-etal-2020-learn} collects multiple human responses to estimate human disagreement for each instance.
This meta-information can potentially help us identify interpretable patterns for the disagreeing instances where one model is better than the other. 
On the flip side, identifying disagreeing instances between two models can also help us  generate hypothesis and decide what subgroup information to annotate.

We can also add performance information on other tasks to each instance.
For example, \citet{pruksachatkun-etal-2020-intermediate} studied the correlation between syntactic probing accuracy \cite{hewitt-liang-2019-designing} and downstream task performance.
\citet{turc2019} and \citet{kaplan2020scaling} studied the correlation between language modelling loss and the downstream task performance.
However, they did not analyze correlations at the instance level.
We may investigate whether their results hold on the instance level:
if an instance is easier to tag by a probe or easier to predict by a larger language model, is the accuracy likely to be higher?

\paragraph{Statistical Tools}
Data mining is more powerful with better statistical tools. 
Initially we used the Benjamini-Hochberg procedure with Fisher's exact test, which required us to pretrain 10 models to formally verify that the decaying instances exist.
However, we later realized that 2 is in fact enough by using our approach introduced in Section \ref{sec:formal-guarantee}. 
We could have saved 80\% of the computation for pretraining if this approach was known before we started. 

Future work can explore more complicated metrics and settings. 
We compared at most 3 different model sizes at a time, and higher order comparisons require novel metrics. 
We studied two sources of randomness, pretraining and finetuning, but other sources of variation can be interesting as well, e.g. differences in pretraining corpus, different model checkpoints, etc. 
To deal with more sophisticated metrics, handle different sources and hierarchies of randomness, and reach conclusions that are robust to noises at the instance level, researchers need to develop new inference procedures.

To conclude, for better instance level understanding, we need to produce and share more prediction data, annotate more diverse linguistic properties, and develop better statistical tools to infer under noises. 
We hope our work can inform researchers about the core challenges underlying instance level understanding and inspire future work.

\section*{Acknowledgement}
We thank Steven Cao, Cathy Chen, Frances Ding, David Gaddy, Colin Li, and Alex Wei for giving comments on the initial paper draft. 
We would also like to thank the Google Cloud TPU team for their hardware support.

\bibliographystyle{acl_natbib}
\bibliography{anthology,acl2021}

\appendix
\newpage
\phantom{0}
\newpage
\section{Pretraining and Finetuning Details} \label{sec:app-prediction-data}
Here we explain how to obtain the model predictions, which are analyzed in later sections.
To obtain these predictions under the ``pretraining and finetuning" framework \cite{devlin-etal-2019-bert}, we need to decide a model \textbf{size}, perform \textbf{pretraining}, finetune on a \textbf{training set} with a choice of \textbf{hyper-parameters}, and test the model on an \textbf{evaluation set}. 
We discuss each bolded aspects below.

\paragraph{Size} 
Similar to \citet{turc2019}, we experimented with the following five model sizes, listed in increasing order: \minisize (L4/H256)
\footnote{4 Layers with hidden dimension 256}
, \smallsize (L4/H512), \mediumsize (L8/H512), \basesize  (L12/H768), and \largesize (L24/H1024).

\paragraph{Pretraining}
We used the pretraining code from \citet{devlin-etal-2019-bert} and the pre-training corpus from \citet{Li2020Efficient}.
Compared to the original BERT release, we used context size 128 instead of 512, since computation cost grows quadratically with respect to context size;
we also pretrained for 2M steps instead of 1M.

\paragraph{Training Set}
We consider 3 datasets:  Quora Question Pairs (QQP)
\footnote{https://www.quora.com/q/quoradata/First-Quora-Dataset-Release-Question-Pairs}, Multi-Genre Natural Language Inference (MNLI; \citet{williams-etal-2020-predicting}), and the Stanford Sentiment Treebank (SST-2; \cite{socher-etal-2013-recursive}).
For QQP we used the official training split. 
For MNLI we used 350K out of 400K instances from the original training split, and added the remaining 50K to the evaluation set, since the original in-domain development set only contains 10K examples.
For SST-2, we mix the training and development set of the original split, split the instances into 5 folds, train on four of them, and evaluate on the remaining fold. 

\paragraph{Hyperparameters}
As in \citet{turc2019}, we finetune 4 epochs for each dataset. 
For each task and model size, we tune hyperparameters in the following way: we first randomly split our new training set into 80\% and 20\%; then we finetune on the 80\% split with all 9 combination of batch size [16, 32, 64] and learning rate [1e-4, 5e-5, 3e-5], and choose the combination that leads to the best average accuracy on the remaining 20\%. 

\paragraph{Evaluation Set}
After finetuning our pretrained models, we evaluate them on a range of in-domain, out-of-domain, or challenging datasets to obtain model predictions. 
Models trained on MNLI are also evaluated on Stanford Natural Language Inference (SNLI; \cite{bowman-etal-2015-large}), Heuristic Analysis for NLI Systems (HANS; \cite{mccoy-etal-2019-right}), and stress test evaluations (STRESS; \cite{naik-etal-2018-stress}).
Models trained on QQP are also evaluated on Twitter Paraphrase Database (TwitterPPDB; \cite{lan-etal-2017-continuously}).

Since pretraining introduces randomness, for each model size $s$, we pretrain 10 times with different random seed $P$;
since finetuning also introduces noise, for each pretrained model we pretrain 5 times with different random seed $F$;
besides, we also evaluate the model at the checkpoints after $E$ epochs, where $E \in [3, 3\frac{1}{3}, 3\frac{2}{3}, 4]$.

Pretraining 10 models for all 5 model sizes altogether takes around 3840 hours on TPU v3 with 8 cores. 
Finetuning all of them 5 times for all three tasks in our paper requires around 1200 hours. 

\section{Compare Our Models to the Original} \label{sec:compare-to-orig}
Since we decreased the pre-training context length to save computation, these models are not exactly the same as the original BERT release by \citet{devlin-etal-2019-bert} and \citet{turc2019}. 
We need to benchmark our model against theirs to ensure that the performance of our model is still reasonable and the qualitative trend still holds.
For each each size and task, we finetune the original model 5 times and calculate the average of overall accuracy. 

The comparison can be seen in Table \ref{tab:compare-to-orig}. 
We find that our model does not substantially differ from the original ones on QQP and SST-2. 
On MNLI, the performance of our BERT-\basesize and BERT-\largesize is 2$\sim$3\% below the original release, but the qualitative trend that larger models have better accuracy still holds robustly. 

\begin{table}[]
    \centering
    \begin{tabular}{lrrr}
    \hline
     &   QQP &  MNLI &  SST-2 \\
    \hline
    \minisize$^{\text{orig}}$   & 88.2\% & 74.6\% &  92.8\% \\
    \minisize$^{\text{ours}}$   & 87.3\% & 74.3\% &  92.8\% \\
    \hline
    \smallsize$^{\text{orig}}$  & 89.1\% & 77.3\% &  93.9\% \\
    \smallsize$^{\text{ours}}$  & 88.7\% & 76.7\% &  93.9\% \\
    \hline
    \mediumsize$^{\text{orig}}$ & 89.8\% & 79.6\% &  94.2\% \\
    \mediumsize$^{\text{ours}}$ & 89.5\% & 78.9\% &  94.2\% \\
    \hline
    \basesize$^{\text{orig}}$   & 90.8\% & 83.8\% &  95.0\% \\
    \basesize$^{\text{ours}}$   & 90.6\% & 81.2\% &  94.6\% \\
    \hline
    \largesize$^{\text{orig}}$  & 91.3\% & 86.8\% &  95.2\% \\
    \largesize$^{\text{ours}}$  & 91.0\% & 83.8\% &  94.8\% \\
    \hline
    \end{tabular}
    \caption{Comparing our pretrained model (superscript $^{orig}$) to the original release by \citet{devlin-etal-2019-bert} and \citet{turc2019} (superscript $^{ours}$). 
    All pretrained models are finetuned with the training set and tested on the in-distribution evaluation set described in Appendix \ref{sec:app-prediction-data}.}
    \label{tab:compare-to-orig}
\end{table}

\section{More Instance Difference Results} \label{app:more-instance-cmp}
\begin{figure*}[t]%
    \centering
    \subfloat[\centering]{{\includegraphics[width=0.4\columnwidth]{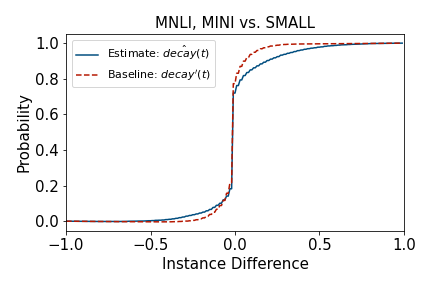} }}%
    \space
    \subfloat[\centering]{{\includegraphics[width=0.4\columnwidth]{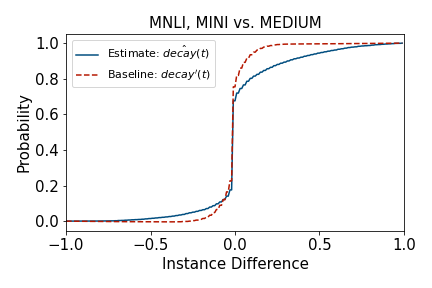} }}%
    \subfloat[\centering]{{\includegraphics[width=0.4\columnwidth]{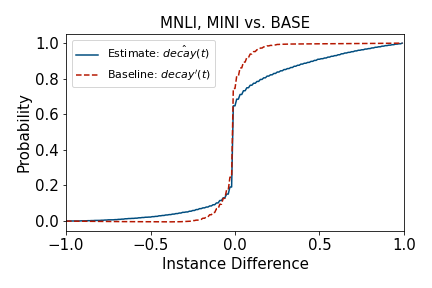} }}%
    \space
    \subfloat[\centering]{{\includegraphics[width=0.4\columnwidth]{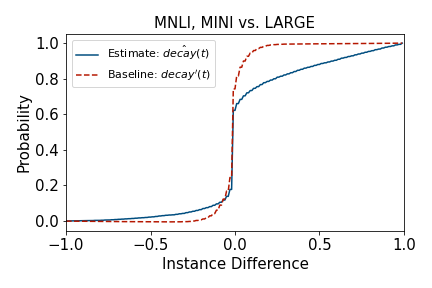} }}%
    \space
    \subfloat[\centering]{{\includegraphics[width=0.4\columnwidth]{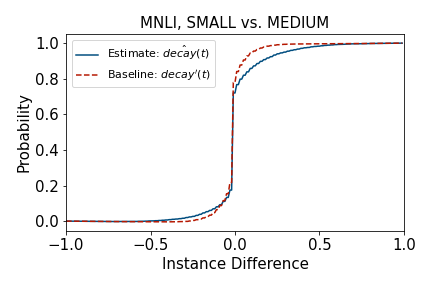} }}%
    \quad
    \subfloat[\centering]{{\includegraphics[width=0.4\columnwidth]{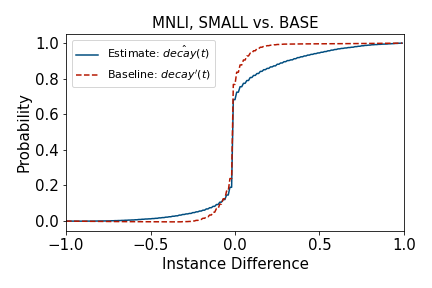} }}%
    \space
    \subfloat[\centering]{{\includegraphics[width=0.4\columnwidth]{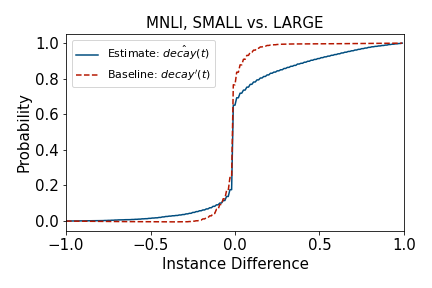} }}%
    \subfloat[\centering]{{\includegraphics[width=0.4\columnwidth]{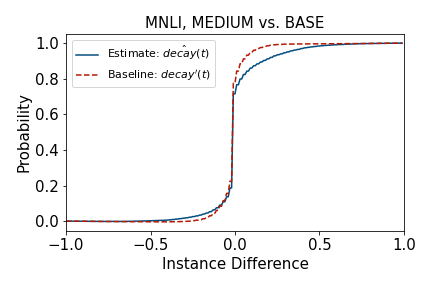} }}%
    \space
    \subfloat[\centering]{{\includegraphics[width=0.4\columnwidth]{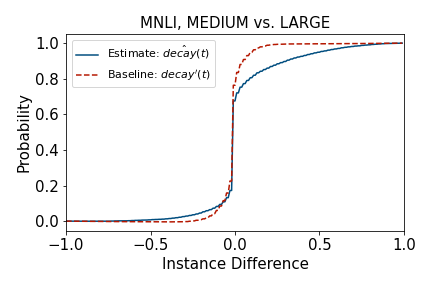} }}%
    \space
    \subfloat[\centering]{{\includegraphics[width=0.4\columnwidth]{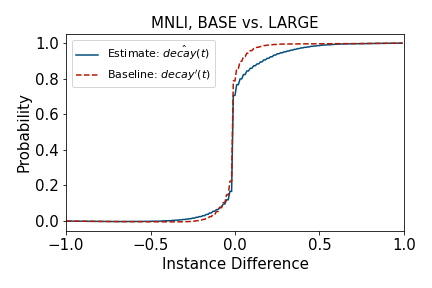} }}%
    \caption{Similar to Figure \ref{fig:cum_instance_diff}, on MNLI in-distribution development set, for each pair of model sizes, we plot the cumulative density function of instance differences.}%
    \label{fig:cum_mnli_all}
\end{figure*}

\begin{figure*}[h]%
    \centering
    \subfloat[\centering]{{\includegraphics[width=0.4\columnwidth]{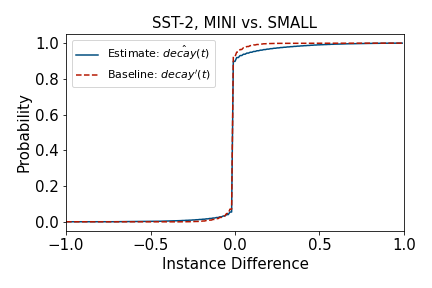} }}%
    \space
    \subfloat[\centering]{{\includegraphics[width=0.4\columnwidth]{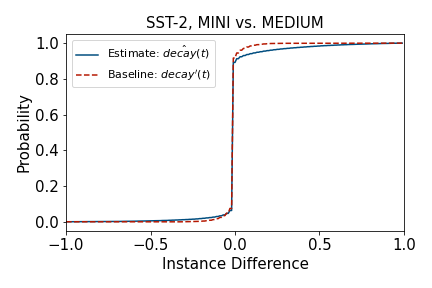} }}%
    \subfloat[\centering]{{\includegraphics[width=0.4\columnwidth]{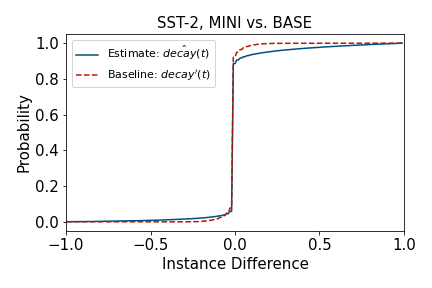} }}%
    \space
    \subfloat[\centering]{{\includegraphics[width=0.4\columnwidth]{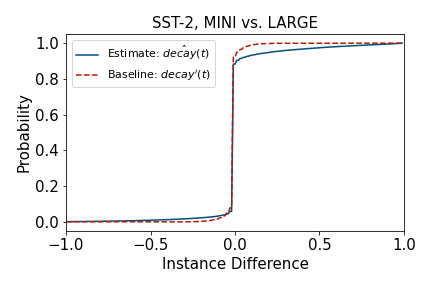} }}%
    \space
    \subfloat[\centering]{{\includegraphics[width=0.4\columnwidth]{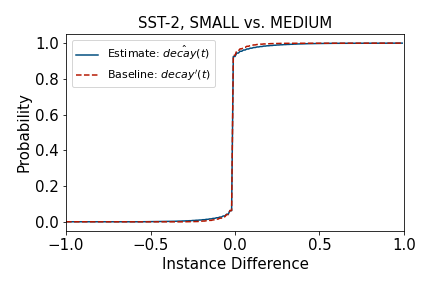} }}%
    \quad
    \subfloat[\centering]{{\includegraphics[width=0.4\columnwidth]{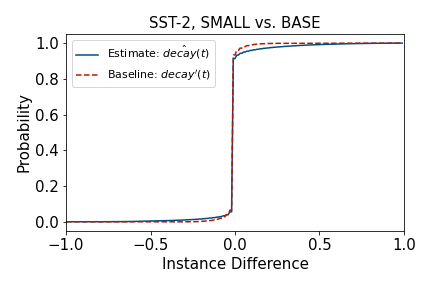} }}%
    \space
    \subfloat[\centering]{{\includegraphics[width=0.4\columnwidth]{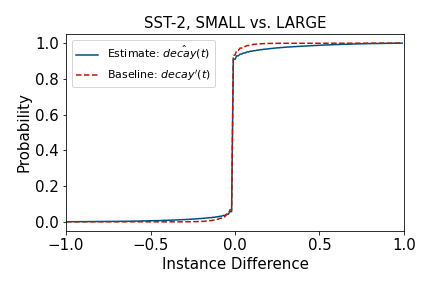} }}%
    \subfloat[\centering]{{\includegraphics[width=0.4\columnwidth]{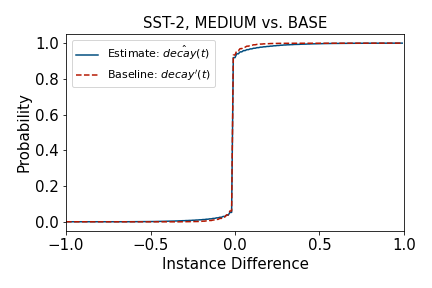} }}%
    \space
    \subfloat[\centering]{{\includegraphics[width=0.4\columnwidth]{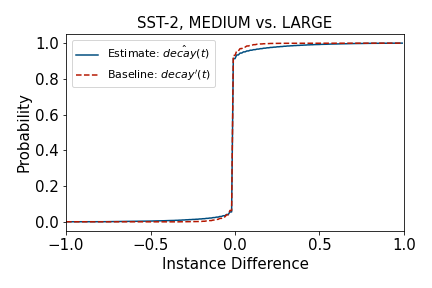} }}%
    \space
    \subfloat[\centering]{{\includegraphics[width=0.4\columnwidth]{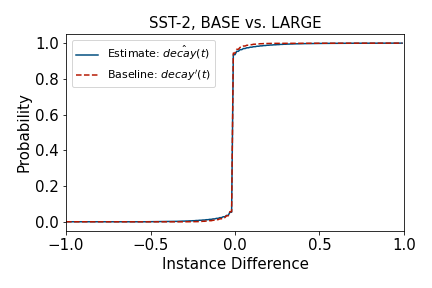} }}%
    \caption{Similar to Figure \ref{fig:cum_instance_diff}, on SST-2, for each pair of model sizes, we plot the cumulative density function of instance differences.}%
    \label{fig:cum_sst-2_all}
\end{figure*}

\begin{figure*}[h]%
    \centering
    \subfloat[\centering]{{\includegraphics[width=0.4\columnwidth]{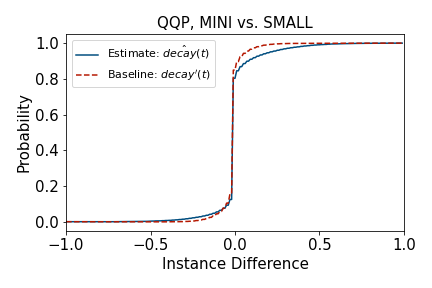} }}%
    \space
    \subfloat[\centering]{{\includegraphics[width=0.4\columnwidth]{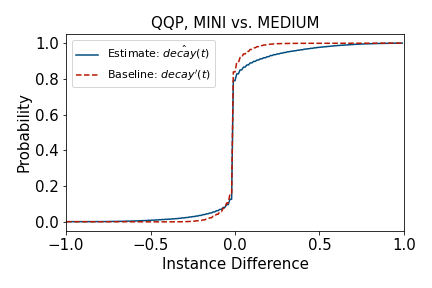} }}%
    \subfloat[\centering]{{\includegraphics[width=0.4\columnwidth]{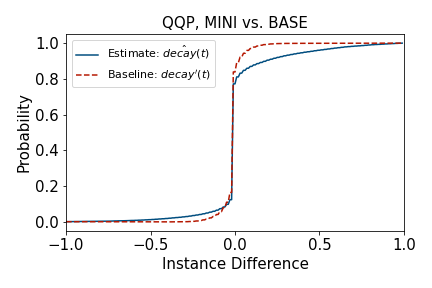} }}%
    \space
    \subfloat[\centering]{{\includegraphics[width=0.4\columnwidth]{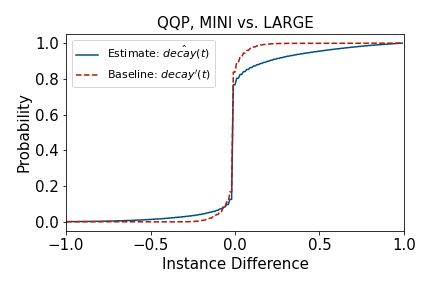} }}%
    \space
    \subfloat[\centering]{{\includegraphics[width=0.4\columnwidth]{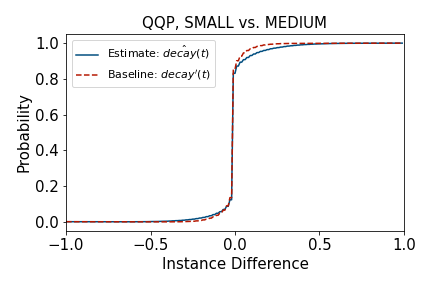} }}%
    \quad
    \subfloat[\centering]{{\includegraphics[width=0.4\columnwidth]{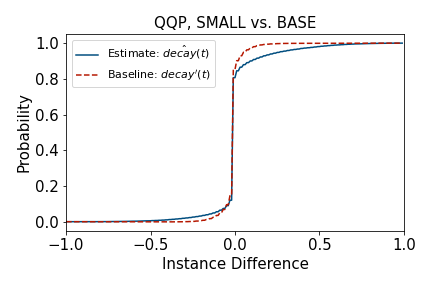} }}%
    \space
    \subfloat[\centering]{{\includegraphics[width=0.4\columnwidth]{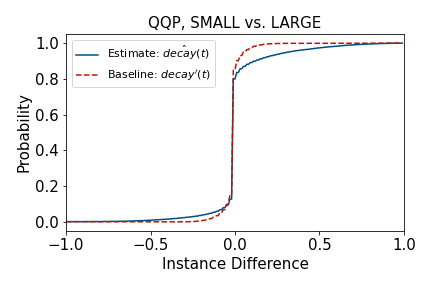} }}%
    \subfloat[\centering]{{\includegraphics[width=0.4\columnwidth]{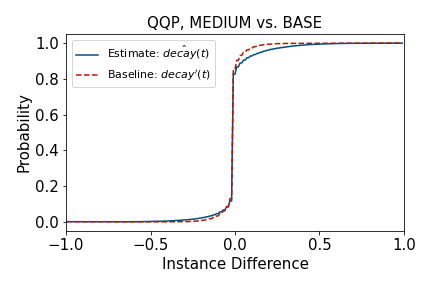} }}%
    \space
    \subfloat[\centering]{{\includegraphics[width=0.4\columnwidth]{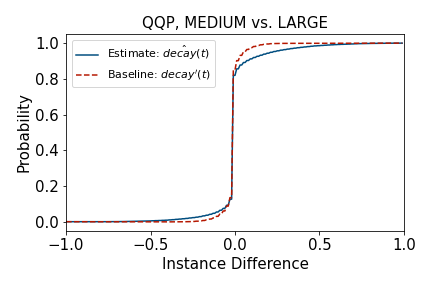} }}%
    \space
    \subfloat[\centering]{{\includegraphics[width=0.4\columnwidth]{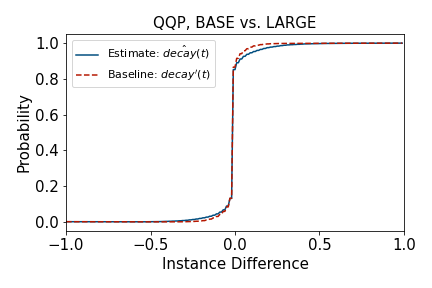} }}%
    \caption{Similar to Figure \ref{fig:cum_instance_diff}, on QQP in-domain development set, for each pair of model sizes, we plot the cumulative density function of instance differences.}%
    \label{fig:cum_qqp_all}
\end{figure*}

Similar to Figure \ref{fig:cum_instance_diff}, for all 10 pairs of model sizes and all in-distribution instances of MNLI, SST-2, and QQP, we plot the cumulative density of $\hat{\Delta \acc}$ and $\Delta \acc'$, or say, $\hat{\decay}(t)$ and $\decay'(t)$ in Figure \ref{fig:cum_mnli_all}, \ref{fig:cum_sst-2_all}, and \ref{fig:cum_qqp_all}.

Additionally, for each pair of model sizes $s_{1}$ and $s_{2}$, we estimate ``how much instances are getting better/worse accuracy?" by taking the maximum difference between the red curve and the blue curve. 
We report these results for MNLI, SST-2, and QQP in Table \ref{tab:all-cmp}. 
We find that larger model size gaps lead to larger decaying fraction, but also larger improving fraction as well. 

\begin{table*}[]
    \centering
    \begin{tabular}{l|rrrrr}
        \hline
        MNLI\phantom{-2} $s_{1}\setminus s_{2}$ &  \minisize &  \smallsize &  \mediumsize &  \basesize &  \largesize \\
        \hline
        \minisize  &   0.000 &  0.087 &   0.136 & 0.179 &  0.214 \\
        \smallsize   & 0.033 &    0.000 &   0.089 & 0.139 &  0.180 \\
        \mediumsize & 0.050 &  0.028 &     0.000 & 0.090 &  0.143 \\
        \basesize   & 0.060 &  0.048 &   0.026 &   0.000 &  0.101 \\
        \largesize  & 0.059 &  0.052 &   0.040 & 0.021 &    0.000 \\
    \end{tabular}

    \begin{tabular}{l|rrrrr}
        \hline
        QQP\phantom{-00} $s_{1}\setminus s_{2}$ &  \minisize &  \smallsize &  \mediumsize &  \basesize &  \largesize \\
        \hline
        \minisize  &   0.000 &  0.057 &   0.076 & 0.100 &  0.107 \\
        \smallsize   & 0.019 &    0.000 &   0.039 & 0.073 &  0.084 \\
        \mediumsize & 0.029 &  0.014 &     0.000 & 0.044 &  0.063 \\
        \basesize   & 0.034 &  0.027 &   0.016 &   0.000 &  0.032 \\
        \largesize  & 0.036 &  0.031 &   0.027 & 0.016 &    0.000 \\
    \end{tabular}
    
    \begin{tabular}{l|rrrrr}
        \hline
        SST-2\phantom{M} $s_{1}\setminus s_{2}$ &  \minisize &  \smallsize &  \mediumsize &  \basesize &  \largesize \\
        \hline
        \minisize   &   0.000 &  0.037 &   0.043 & 0.052 &  0.057 \\
        \smallsize  & 0.010 &    0.000 &   0.015 & 0.031 &  0.036 \\
        \mediumsize & 0.016 &  0.008 &     0.000 & 0.020 &  0.028 \\
        \basesize  & 0.019 &  0.014 &   0.009 &   0.000 &  0.014 \\
        \largesize  & 0.020 &  0.017 &   0.015 & 0.008 &    0.000 \\
        \hline
    \end{tabular}

    \caption{On QQP, MNLI in domain development set and SST-2 we lowerbound the fraction of instances that improves when model size changes from $s_{1}$ (row) to $s_{2}$ (column). }
    \label{tab:all-cmp}
\end{table*}

\section{Proof of Theorem \ref{thm:decaylowerbound}} \label{sec:our-fdr-estimate}
\paragraph{Formal Setup}
Our goal is to show that if all the random seeds are independent, 

\begin{equation}
    \decay \geq \mathbb{E}[\hat{\decay}(t) - \decay'(t)]
\end{equation}

More concretely, suppose each instance is indexed by $i$, the set of all instances is $\mathcal{T}$, and the random seed is $R$; then $c^{s}_{R} \in \{0, 1\}^{|\mathcal{T}|}$ is a random $|\mathcal{T}|$ dimensional vector, where $c^{s}_{R, i} = 1$ if the model of size $s$ correctly predicts instance $i$ under the random seed $R$.
We are comparing model size $s_{1}$ and $s_{2}$, where $s_{2}$ is larger; to keep notation uncluttered, we omit these indexes whenever possible.

Suppose we observe $c^{s_{1}}_{R_{1\dots 2k}}$ and $c^{s_{2}}_{R_{2k+1\dots 4k}}$, where there are $2k$ different random seeds for each model size \footnote{We assumed even number of random seeds since we will mix half of the models from each size to compute the random baseline}. Then
\begin{equation} \label{eq:estimate-concrete}
    \hat{\Delta \acc}_{i} := \frac{1}{2k}(\sum^{2k}_{j=1}c^{s_{1}}_{R_{j},i} - \sum^{4k}_{j=2k+1}c^{s_{2}}_{R_{j},i}),
\end{equation}

and hence the discovery rate $\hat{\decay}(t)$ is defined as 
\begin{align}
    \hat{\decay}(t) := \frac{1}{|\mathcal{T}|}\sum_{i=1}^{|\mathcal{T}|}\mathbf{1}[\hat{\Delta \acc} \leq t].
\end{align}

For the random baseline estimator, we have
\begin{align}\label{eq:control-concrete}
    \Delta \acc'_{i} &:= \frac{1}{2k}(\sum^{k}_{j=1}c^{s_{1}}_{R_{j}, i} + \sum^{3k}_{j=2k+1}c^{s_{2}}_{R_{j}, i} \\
    &- \sum^{2k}_{j=k+1}c^{s_{1}}_{R_{j},i} - \sum^{4k}_{j=3k+1}c^{s_{2}}_{R_{j},i} \nonumber
    ),
\end{align}

and the false discovery control $\decay'$ is defined as 
\begin{align}
    \decay'(t) := \frac{1}{|\mathcal{T}|}\sum_{i=1}^{|\mathcal{T}|}\mathbf{1}[\Delta \acc'_{i} \leq t].
\end{align}

To reiterate, the definition of the true decay rate is  
\begin{equation}
    \decay = \frac{1}{|\mathcal{T}|}\sum_{i=1}^{|\mathcal{T}|}\mathbf{1}[\Delta\acc_{i} < 0].
\end{equation}

Our goal is to prove that 
\begin{equation} \label{eq:target-decay-bound}
    \decay \geq \mathbb{E}_{R_{1}\dots R_{4k}}[\hat{\decay}(t) - \decay'(t)]
\end{equation}

\paragraph{Proof} 
By re-arranging terms and linearity of expectation, Equation \ref{eq:target-decay-bound} is equivalent to the following
\begin{align}
    \sum_{i=1}^{|\mathcal{T}|}(\mathbf{1}[\Delta \acc_{i} < 0] - \mathbb{P}[\hat{\Delta \acc}_{i} \leq t] \\+ \mathbb{P}[\Delta \acc'_{i} \leq t]) \geq 0 \nonumber
\end{align}

Hence, we can declare victory if we can prove that for all $i$, 
\begin{align} \label{eq:instance-target}
    \mathbf{1}[\Delta \acc_{i} < 0] - \mathbb{P}[\hat{\Delta \acc}_{i} \leq t] \\+ \mathbb{P}[\Delta \acc'_{i} \leq t] \geq 0\nonumber
\end{align}

To prove Equation \ref{eq:instance-target}, we observe that if $\acc_{i} < 0$, since the probabilities are bounded by 0 and 1, its left-hand side must be positive. 
Therefore, we only need to prove that
\begin{align} \label{eq:instance-lowest-target}
    &\Delta \acc_{i} \geq 0 \\
    &\Rightarrow \mathbb{P}[\Delta \acc'_{i} \leq t] \geq \mathbb{P}[\hat{\Delta \acc}_{i} \leq t], \nonumber
\end{align}
which will be proved in Lemma \ref{thm:fdr-lemma}.$\square$

\begin{lemma} \label{thm:fdr-lemma}
\begin{align} \label{eq:instance-lowest-target}
    &\Delta \acc_{i} \geq 0 \\
    &\Rightarrow \mathbb{P}[\Delta \acc'_{i} \leq t] \geq \mathbb{P}[\hat{\Delta \acc}_{i} \leq t], \nonumber
\end{align}
\end{lemma}
For $m = 1, 2$, define
\begin{equation}
    p^{s_{m}}_{i} := \mathbb{E}_{R}[c^{s_{m}}_{i}],
\end{equation}
then
\begin{equation}
    p^{s_{1}}_{i} \leq p^{s_{2}}_{i}
\end{equation}

Since $c^{s_{1}}_{i}$ and $c^{s_{2}}_{i}$ are both Bernoulli random variables with rate $p^{s_{1}}_{i}$ and $p^{s_{2}}_{i}$ respectively, we can write down the probability distribution of $\hat{\Delta \acc}_{i}$ and $\Delta \acc'_{i}$ as the sum/difference of several binomial variables, i.e.

\begin{align}
    &\hat{\Delta\acc}_{i} \sim (\text{Binom}(k, p^{s_{2}}_{i}) + \text{Binom}(k, p^{s_{2}}_{i}) \\
    &- \text{Binom}(k, p^{s_{1}}_{i}) - \text{Binom}(k, p^{s_{1}}_{i})) / 2k, \nonumber
\end{align}

and 
\begin{align}
    &\Delta\acc'^{i} \sim (\text{Binom}(k, p^{s_{2},i}) + \text{Binom}(k, p^{s_{1},i}) \\
    &- \text{Binom}(k, p^{s_{1},i}) - \text{Binom}(k, p^{s_{2},i})) / 2k \nonumber
\end{align}

$p^{s_{1}}_{i} \leq p^{s_{2}}_{i}$, $\text{Binom}(k, p^{s_{2},i}))$ first order stochastically dominates $\text{Binom}(k, p^{s_{1},i})$. 
Therefore, $\Delta\acc'^{i}$ dominates $\hat{\Delta\acc}_{i}$, hence completing the proof. $\square$

\subsection{Independent Seed Assumption} \label{sec:independent-seed-assumption}
We notice that Theorem \ref{thm:decaylowerbound} requires the seeds $R$ to be independent.
This assumption does not hold on our data, since some finetuning runs share the same pretraining seeds.
Therefore, the above proof no longer holds. 
Specifically, Lemma \ref{thm:fdr-lemma} fails because $\hat{\Delta\acc}$ and $\Delta\acc'$ are no longer binomial variables, and the later does not necessarily dominate the first.
Here is a counter-example, if the seeds are not entirely independent.

Hypothetically, suppose we are comparing a smaller model $s_{1}$ and a larger model $s_{2}$. 
For the smaller model, with .1 probability it finds a perfect pretrained model that always predict correctly across all finetuning runs and with .9 probability it finds a bad pretrained model that predict always incorrectly.
For the larger model, with probability 1 it finds an average pretrained model that predict correctly for .2 fraction of finetuning runs.
The larger model is on average better, because it has $.2 > .1$ probability to be correct.
Hence, $\Delta \acc > 0$

Suppose we observe 2 independent pretraining seeds for each size and infinite number of finetuning seeds for each pretraining seed, and let us consider the threshold -0.8. Then
\begin{align}
    \mathbb{P}[\hat{\Delta \acc}_{i} \leq -0.8] \\= 0.01 \geq 0 = \mathbb{P}[\Delta \acc'_{i} \leq -0.8] 
\end{align}
The event that $\hat{\Delta \acc}_{i} \leq -0.8$ happens with probability 0.01 when both of the two small pretrained models have good pretraining seeds, and $\Delta \acc'_{i}$ is at least -0.5 and will never be less than -0.8.

The key idea behind this counter-example is that even if the larger model has better average, the distribution of average finetuning accuracy for different pretraining seeds might not stochastically dominate the one with lower average because of outliers. 
Hence, a priori, this is unlikely to happen in practice, since pretraining variance is generally small, and we have multiple pretraining seeds to average out the outliers.
Nevertheless, future work is needed to make a more rigorous argument.

\section{Upward Bias of Adaptive Thresholds} \label{sec:adaptive-thresholds}
In section \ref{sec:instance-acc-cmp} we picked the best threshold that can maximize the lowerbound, which can incur a slight upward bias. 
Here we estimate that the bias is at most 10\% relative to the unbiased lowerbound with a bootstrapping method. 

We use the empirical distribution of 10 pretrained models as the ground truth distribution for bootstrapping. 
We first compute a best threshold with 10 sampled smaller and larger pretrained models, and then compute the lowerbound $L$ with this threshold on another sample of 10 smaller and larger models. 
Intuitively, we use one bootstrap sample (which contains 10 smaller pretrained models and 10 larger pretrained models) as the development set to ``tune the threshold", and then use this threshold on a fresh bootstrap sample to compute the lowerbound.
We refer to the lowerbound that uses the best threshold as $L^{*}$, and compute the relative error $\mathbb{E}[(L^{*} - L)] / \mathbb{E}[L)]$, where the expectation is taken with respect to bootstrap samples.

We report all results in Table \ref{tab:all-cmp-upwardbias}.
In general, we find that the upward bias is negligible, which is at most around $10\%$.

\begin{table*}[]
    \centering
    \begin{tabular}{l|rrrrr}
        \hline
        MNLI\phantom{-2} $s_{1}\setminus s_{2}$ &  \minisize &  \smallsize &  \mediumsize &  \basesize &  \largesize \\
        \hline
        \minisize   & 0.000 &  0.031 &   0.027 & 0.026 &  0.020 \\
        \smallsize  & 0.108 &  0.000 &   0.027 & 0.023 &  0.019 \\
        \mediumsize & 0.095 &  0.116 &   0.000 & 0.028 &  0.023 \\
        \basesize   & 0.093 &  0.100 &   0.144 & 0.000 &  0.026 \\
        \largesize  & 0.097 &  0.103 &   0.117 & 0.149 &  0.000 \\
    \end{tabular}

    \begin{tabular}{l|rrrrr}
        \hline
        QQP\phantom{-00} $s_{1}\setminus s_{2}$ &  \minisize &  \smallsize &  \mediumsize &  \basesize &  \largesize \\
        \hline
        \minisize   & 0.000 &  0.025 &   0.022 & 0.021 &  0.020 \\
        \smallsize  & 0.127 &  0.000 &   0.040 & 0.020 &  0.020 \\
        \mediumsize & 0.093 &  0.146 &   0.000 & 0.032 &  0.031 \\
        \basesize   & 0.087 &  0.119 &   0.123 & 0.000 &  0.049 \\
        \largesize  & 0.090 &  0.105 &   0.079 & 0.106 &  0.000 \\
    \end{tabular}
    
    \begin{tabular}{l|rrrrr}
        \hline
        SST-2\phantom{M} $s_{1}\setminus s_{2}$ &  \minisize &  \smallsize &  \mediumsize &  \basesize &  \largesize \\
        \hline
        \minisize   & 0.000 &  0.028 &   0.022 & 0.021 &  0.019 \\
        \smallsize  & 0.117 &  0.000 &   0.047 & 0.031 &  0.029 \\
        \mediumsize & 0.075 &  0.093 &   0.000 & 0.063 &  0.035 \\
        \basesize   & 0.068 &  0.067 &   0.085 & 0.000 &  0.071 \\
        \largesize  & 0.071 &  0.067 &   0.060 & 0.098 &  0.000 \\
        \hline
    \end{tabular}
    
    \caption{The same table as \ref{tab:all-cmp-upwardbias}, except that we are now calculating the relative upward bias $\mathbb{E}[(L^{*} - L)] / \mathbb{E}[L)]$ as described in Section \ref{sec:adaptive-thresholds}.}
    \label{tab:all-cmp-upwardbias}
\end{table*}

\section{Comparison with Significance Testing} \label{sec:cmp-signifiance}
We also experimented with the classical approach that calculates the significance-level for each instance and then use the Benjamini-Hochberg procedure to lowerbound the decaying fraction.
To make sure that we are comparing with this approach fairly, we lend it additional power by picking the false discovery rate that can maximize the true discovery counts.
We report the decaying fraction on MNLI in-domain development set found by this classical method and compare it with our method for different model size differences in Table \ref{tab:compare-ours-bh}; we also simulate situations when we have fewer models.

In general, we find that our method always provide a tighter (higher) lowerbound than the classical method, and 2 models are sufficient to verify the existence (i.e. lowerbound $>$ 0) of the decaying fraction; 
in contrast, the classical method sometimes fails to do this even with 10 models, e.g., when comparing \basesize to \largesize.

Intuitively, our approach provides a better lowerbound because it better makes use of the information that on most instances, both the smaller and the larger models agree and predict completely correctly or incorrectly\footnote{This is for intuition, though, and we do not need any assumption on the prior of instance accuracy, which requires a Bayes interpretation.}. 
For an extreme example, suppose  we only observe 2 smaller models and 2 larger models, and infinite number of datapoints, whose predictions are independent. 
On 99.98\% datapoints, both models have instance accuracy 1; on 0.01\% datapoints, smaller model is completely correct while bigger completely wrong, while on the rest 0.01\% smaller completely wrong but bigger completely correct. 
Setting threshold to be 2, our decay estimate $\hat{\decay}$ is 0.01\%, while $\decay' = 0$: since the models either completely predict correct or wrongly, there is never a false discovery. 
Therefore, our method can provide the tightest lowerbound 0.01\% in this case.
On the other hand, since we only have 4 models in total, the lowest significance-level given by the fisher exact test is 17\% $\gg 0.1\%$, hence the discovery made by the Benjamin-Hochberg procedure is 0.

\begin{table*}[]
    \centering
    \begin{tabular}{lllrrrrr}
    \hline
    s1 &      s2 & method &     2 &     4 &     6 &     8 &    10 \\
    \hline
     \minisize &   \smallsize &   ours & 0.004 & 0.011 & 0.016 & 0.020 & 0.023 \\
    \minisize &   \smallsize &     BH & 0.000 & 0.000 & 0.000 & 0.000 & 0.002 \\
    \hline
    \minisize &  \mediumsize &   ours & 0.012 & 0.019 & 0.026 & 0.032 & 0.035 \\
    \minisize &  \mediumsize &     BH & 0.000 & 0.000 & 0.003 & 0.008 & 0.011 \\
    \hline
    \minisize &    \basesize &   ours & 0.019 & 0.028 & 0.035 & 0.040 & 0.042 \\
    \minisize &    \basesize &     BH & 0.000 & 0.000 & 0.008 & 0.014 & 0.020 \\
    \hline
    \minisize &   \largesize &   ours & 0.019 & 0.027 & 0.031 & 0.037 & 0.040 \\
    \minisize &   \largesize &     BH & 0.000 & 0.000 & 0.009 & 0.015 & 0.019 \\
    \hline
    \smallsize &  \mediumsize &   ours & 0.002 & 0.006 & 0.010 & 0.015 & 0.017 \\
    \smallsize &  \mediumsize &     BH & 0.000 & 0.000 & 0.000 & 0.000 & 0.000 \\
    \hline
    \smallsize &    \basesize &   ours & 0.013 & 0.020 & 0.025 & 0.030 & 0.033 \\
    \smallsize &    \basesize &     BH & 0.000 & 0.000 & 0.002 & 0.006 & 0.011 \\
    \hline
    \smallsize &   \largesize &   ours & 0.015 & 0.021 & 0.026 & 0.031 & 0.033 \\
    \smallsize &   \largesize &     BH & 0.000 & 0.000 & 0.005 & 0.009 & 0.013 \\
    \hline
    \mediumsize &    \basesize &   ours & 0.006 & 0.010 & 0.013 & 0.014 & 0.016 \\
    \mediumsize &    \basesize &     BH & 0.000 & 0.000 & 0.000 & 0.000 & 0.001 \\
    \hline
    \mediumsize &   \largesize &   ours & 0.010 & 0.014 & 0.019 & 0.022 & 0.023 \\
    \mediumsize &   \largesize &     BH & 0.000 & 0.000 & 0.002 & 0.004 & 0.006 \\
    \hline
    \basesize &   \largesize &   ours & 0.004 & 0.005 & 0.009 & 0.010 & 0.012 \\
    \basesize &   \largesize &     BH & 0.000 & 0.000 & 0.000 & 0.000 & 0.000 \\
    \hline
    \end{tabular}
    \caption{We compare each pair of model sizes $s_{1}$ and $s_{2}$ and report the lower bound provided by our method and the Benjamin-Hochberg (BH) procedure. 
    The numbers in column name denote how many pretrained model we used to obtain the lower bounds. 
    }
    \label{tab:compare-ours-bh}
\end{table*}

\section{More Results on Momentum} \label{sec:more-on-momentum}
We report more results on the correlation between instance differences.
Specifically, for one triplet of model sizes (e.g. \minisize $\Rightarrow$ \mediumsize $\Rightarrow$ \largesize), for each group of instances that have similar $\hat{\acc}^{\mediumsize}$, we calculate the correlation between instance differences, i.e. the Pearson-R score between $^{\minisize}_{\mediumsize}\Delta\acc$ and $^{\mediumsize}_{\largesize}\Delta\acc$.
All results can be seen in Table \ref{tab:all_momentum}.

We observe that
\begin{itemize}
    \item For nearly all buckets, the improvements are positively correlated.
    \item When model size gap becomes larger (e.g. $\minisize,\mediumsize,\largesize$ has the largest model size differences), the correlation decreases.
\end{itemize}

\begin{table*}[]
    \centering
    \begin{tabular}{l|rrrrrrrrrr}
    \hline
    MNLI.\phantom{-} \quad Buckets$\Rightarrow$ & 0.10 & 0.20 & 0.30 & 0.40 & 0.50 & 0.60 & 0.70 & 0.80 & 0.90 & 1.00 \\
    \hline
    $\minisize,\smallsize,\mediumsize$& 0.00 &  0.18 &  0.19 &  0.18 &  0.23 &  0.26 &  0.24 &  0.23 &  0.20 &  0.12 \\
    $\smallsize,\mediumsize,\basesize$ & 0.07 &  0.22 &  0.29 &  0.40 &  0.35 &  0.33 &  0.38 &  0.27 &  0.24 &  0.13 \\
    $\mediumsize,\basesize,\largesize$ & 0.05 &  0.09 &  0.17 &  0.33 &  0.20 &  0.30 &  0.12 &  0.13 &  0.16 &  0.09\\
    $\minisize,\mediumsize,\largesize$ & 0.03 &  0.15 &  0.18 &  0.33 &  0.17 &  0.16 &  0.22 &  0.20 &  0.19 &  0.09 \\
    \hline
    \end{tabular}
    
    \begin{tabular}{l|rrrrrrrrrr}
    QQP.\phantom{0-} \quad Buckets$\Rightarrow$ & 0.10 & 0.20 & 0.30 & 0.40 & 0.50 & 0.60 & 0.70 & 0.80 & 0.90 & 1.00 \\
    \hline
    $\minisize,\smallsize,\mediumsize$   &  0.03 &  0.21 &  0.18 &  0.21 &  0.21 &  0.25 &  0.18 &  0.16 &  0.10 &  0.06 \\
    $\smallsize,\mediumsize,\basesize$   &  0.01 &  0.17 &  0.23 &  0.19 &  0.24 &  0.22 &  0.24 &  0.19 &  0.16 &  0.05 \\
    $\mediumsize,\basesize,\largesize$   & -0.02 &  0.16 &  0.09 &  0.23 &  0.17 &  0.10 &  0.14 &  0.14 &  0.09 & -0.01 \\
    $\minisize,\mediumsize,\largesize$   & -0.01 &  0.07 &  0.14 &  0.09 &  0.16 &  0.09 &  0.16 &  0.07 &  0.10 &  0.07 \\
    \hline
    \end{tabular}
    
    \begin{tabular}{l|rrrrrrrrrr}
    SST-2.\phantom{0} \quad Buckets$\Rightarrow$ & 0.10 & 0.20 & 0.30 & 0.40 & 0.50 & 0.60 & 0.70 & 0.80 & 0.90 & 1.00 \\
    \hline
    $\minisize,\smallsize,\mediumsize$ &  0.09 &  0.26 &  0.43 &  0.22 &  0.28 &  0.24 &  0.27 &  0.35 &  0.20 &  0.12 \\
$\smallsize,\mediumsize,\basesize$ & 0.07 &  0.12 &  0.22 &  0.40 &  0.07 &  0.20 &  0.10 &  0.12 &  0.19 &  0.06 \\
$\mediumsize,\basesize,\largesize$  &  0.01 &  0.24 &  0.29 &  0.35 &  0.19 &  0.19 &  0.26 &  0.39 &  0.15 &  0.03 \\
$\minisize,\mediumsize,\largesize$ & 0.01 &  0.17 &  0.11 &  0.41 &  0.04 &  0.29 &  0.16 &  0.21 &  0.15 &  0.07 \\
    \hline
    \end{tabular}
    \caption{Sorted in ascending order, the model sizes are \minisize, \smallsize, \mediumsize, \basesize, and \largesize. The three model sizes listed for each row represents the model size of interest: for example, $\minisize,\mediumsize,\largesize$ means that we are calculating the correlation between $^{\minisize}_{\mediumsize}\Delta\acc$ and $^{\mediumsize}_{\largesize}\Delta\acc$. Each column $t$ represents a bucket that contains instances with middle size accuracy in $[t-0.1,t]$. For example, if the row name is $\minisize,\mediumsize,\largesize$, then the column 0.2 corresponds to a bucket where $\hat{\acc}^{\mediumsize}_{i}$ is between 0.1 and 0.2. 
    We calculate the PearsonR correlation score between  $^{\minisize}_{\mediumsize}\hat{\Delta\acc}$ and $^{\mediumsize}_{\largesize}\hat{\Delta\acc}$ across all instances in the bucket. 
    }
    \label{tab:all_momentum}
\end{table*}

\section{Loss Decomposition and Estimation} \label{sec:decompose}
In this section, under the bias-variance decomposition and total variance decomposition framework, we decompose loss into four components: bias, variance brought by pretraining randomness, by finetuning randomness, and across different checkpoints throughout training. 
We formally define the quantities we want to estimate in Appendix \ref{sec:decompose-formalize}, present an unbiased estimator for these quantities in Appendix \ref{sec:decompose-unbiased}, and show that our method can be generalized to arbitrary number of source of randomness in Appendix \ref{sec:decompoe-general}. 

Specifically, the main paper focused on scenarios with 2 sources of randomness: pretraining and finetuning.
We discuss the case with 3 sources of randomness in the appendix, rather than 2 as in the main paper, because it is easier to understand the general estimation strategy in the case of 3.

\subsection{Formalizing Decomposition} \label{sec:decompose-formalize}

Recall that $P$ is the pretraining seed, $F$ is the finetuning seed, $E$ represents a model checkpoint, $i$ indexes each instance (datapoint).
$c^{s,i}_{P,F,E} = 1$ if the model of size $s$ with pretraining seed $p$ and finetuning seed $F$, and trained for $E$ epochs is correct on datapoint $i$, and 0 otherwise.
Notice that we move the instance index from the subscript to the superscript, since we now use subscript for random seeds, and instance index can be omitted in most of our derivations.

The expected squared loss $\mathcal{L}$ of model $s$ on instance $i$ can then be written as 
\begin{equation}
    \mathcal{L}^{s,i} = \mathbb{E}_{P,F,E}[(1 - c^{s,i}_{P,F,E})^{2}]
\end{equation}

Since we will analyze this term at a datapoint level, we drop the subscript $s$ and $i$ to keep the notation uncluttered.
By the standard bias variance decomposition and total variance decomposition, we decompose the loss $\mathcal{L}$ into four terms:

\begin{align} \label{eq:decomposition}
    \mathcal{L} = & \biassquared + \pretrainvar\\ 
    &+ \finetunevar + \ckptvar.\nonumber
\end{align}

We will walk through the meaning and definition of these four terms one by one.
$\biassquared$ captures how bad is the average prediction, defined as 

\begin{equation}
    \biassquared = (1 - \mathbb{E}_{P,F,E}[c_{P,F,E}])^{2}.
\end{equation}

$\pretrainvar$ captures the variance brought by randomness in pretraining, and is defined as
\begin{equation}
    \pretrainvar = Var_{P}[\mathbb{E}_{F,E}[c_{P,F,E}]].
\end{equation}

Similarly, we define the variance brought by randomness in finetuning $\finetunevar$
\begin{equation}
    \finetunevar = \mathbb{E}_{P}[Var_{F}[\mathbb{E}_{E}[c_{P,F,E}]]],
\end{equation}

and that by fluctuations across checkpoints $e$
\begin{equation}
    \ckptvar = \mathbb{E}_{P,F}[Var_{E}[c_{P,F,E}]].
\end{equation}

\subsection{Unbiased Estimation} \label{sec:decompose-unbiased}

We first describe the data on which we apply our estimator.
Suppose we pretrain $\mathcal{P}$ models with different random seeds, for each of the $\mathcal{P}$ pretrained models we finetune with $\mathcal{F}$ different random seeds, and we evaluate at $\mathcal{E}$ different checkpoints. 
Then $\forall j \in [\mathcal{P}], k \in [\mathcal{F}],l \in [\mathcal{E}]$
\footnote{[$L$] := $\{l: l \in N, l \in [0, L-1]\}$
}
, we observe $P_{j}, F_{jk}, E_{jkl}, $ and $c_{P_{j}, F_{jk}, E_{jkl}}$, where each observed $P,F$ and $E$ are i.i.d. distributed.
Our goal is to estimate from $c$ the four quantities described in the previous section.

\subsubsection{Estimating $\ckptvar$} \label{sec:estimatingckpttunevar}
It is straightforward to estimate $\ckptvar$. 
The estimator $\hat{\ckptvar}$ defined below is unbiased:
\begin{align}
    \hat{\ckptvar}&:= \frac{1}{\mathcal{P}\mathcal{F}}\sum_{j \in [\mathcal{P}], k\in [\mathcal{F}]} \hat{Var}_{E}^{P_{j},F_{jk}},
\end{align}
where 
\begin{equation}
     \hat{Var}_{E}^{P_{j},F_{jk}}:= \frac{1}{\mathcal{E} - 1}\sum_{l\in\mathcal{E}}(c_{P_{j},F_{jk},E_{jkl}} - \Bar{c}_{P_{j},F_{jk}})^{2},
\end{equation}
and
\begin{equation}
    \Bar{c}_{P_{j},F_{jk}}:= \frac{1}{\mathcal{E}}\sum_{l\in[\mathcal{E}]}c_{P_{j},F_{jk},E_{jkl}}.
\end{equation}

$\hat{\ckptvar}$ is unbiased, since $\hat{Var}_{E}^{P_{j},F_{jk}}$ is an unbiased estimation of variance of $c$ with fixed $P_{j}$ and $F_{jk}$, and randomness $E$, i.e.
\begin{align}
    \mathbb{E}_{E_{ij(\cdot)}}&[\hat{Var}_{E}^{P_{j},F_{jk}}] =\\ &Var_{E}[c_{P,F,E}|P=P_{j}, F=F_{jk}] \nonumber.
\end{align}

Therefore, $\forall j \in [\mathcal{P}], k \in [\mathcal{F}]$, we have
\begin{equation}
    \mathbb{E}_{P_{j},F_{jk}}[\hat{Var}_{E}^{P_{j},F_{jk}}] = \ckptvar, 
\end{equation}
and hence by linearity of expectation
\begin{equation}
    \mathbb{E}_{P_{(\cdot)}, F_{(\cdot)}, E_{(\cdot)}} [\hat{\ckptvar}] = \ckptvar.
\end{equation}

\subsubsection{Estimating $\finetunevar$} \label{sec:estimatingfinetunevar}
As before, by linearity of expectation, we can declare victory if we can develop an unbiased estimator for the following quantity and then average across $P_{j}$:
\begin{equation}
    Var_{F}[\mathbb{E}_{E}[c_{P,F,E}]| P = P_{j}],
\end{equation}
which verbally means "variance across different finetuning seeds of the mean of $c$ over different checkpoints $E$, conditioned on the pretraining seed $P_{j}$."

Since $P_{j}$ is fixed for this estimator, we drop the subscripts $P$ to keep notation uncluttered. 
Therefore, we want to estimate
\begin{equation}\label{eq:varf}
    Var_{F} := Var_{F}[\mathbb{E}_{E}[c_{F,E}]]
\end{equation}

A naive solution is to take first take the mean $\Bar{c}_{F_{k}}$ of $c$ for each $F_{k}$, i.e.
\begin{equation}
    \Bar{c}_{F_{jk}} := \frac{1}{\mathcal{E}}\sum_{l\in[\mathcal{E}]}c_{F_{k}, E_{kl}},
\end{equation}
and then calculate the sample variance $\Tilde{Var}_{F}$ of $\Bar{c}$ with respect to $F$:
\begin{equation}
    \Tilde{Var}_{F} := \frac{1}{\mathcal{F} - 1}\sum_{k\in [\mathcal{F}]}(\Bar{c}_{F_{k}} - \Bar{c})^{2},
\end{equation}
where 
\begin{equation}
    \Bar{c} := \frac{1}{\mathcal{F}}\sum_{k\in [\mathcal{F}]}\Bar{c}_{F_{k}}
\end{equation}

However, this would create an upward bias: the empirical mean $\Bar{c}_{F_{jk}}$ is a noisy estimate of the population mean $\mathbb{E}_{E}[c_{F_{jk},E}]$, and hence increases let $\Tilde{Var}_{F}$ over-estimate the variance. 
Imagine a scenario where $Var_{F}$ is in fact 0; however, since $\Bar{c}_{F_{jk}}$ is a noisy estimate, $\Tilde{Var}_{F}$ will sometimes be positive but never below 0.
As a result, $\mathbb{E}[\Tilde{Var}_{F}] > 0$, which is a biased estimator.

We introduce the following general theorem to correct this bias. 

\begin{theorem} \label{bias-variance-thm-1}
Suppose $\mathcal{D}_{k}, k\in [\mathcal{F}]$ are independently sampled from the same distribution $\Xi$, which is a distribution of distributions;
$\hat{\mu}_{k}$ is an unbiased estimator of $\mathbb{E}_{X\in\mathcal{D}_{k}}[X]$, and $\hat{\phi}_{k}$ to be an unbiased estimator of the variance of $\hat{\mu}_{k}$, then

\begin{align} \label{eq:unbiased-variance-across-mean}
    \hat{Var}_{F} &= \frac{1}{\mathcal{F}-1}\sum_{k\in[\mathcal{F}]}(\hat{\mu}_{k}-\hat{\mu})^{2} \\
    &- \frac{1}{\mathcal{F}}\sum_{k\in[\mathcal{F}]}\hat{\phi}_{k} \nonumber
\end{align}

is an unbiased estimator for
\begin{equation}
    V = Var_{\mathcal{D}\sim\Xi}[\mathbb{E}_{X\sim\mathcal{D}}[X]],
\end{equation}

where
\begin{equation}
    \hat{\mu} := \frac{1}{\mathcal{F}}\sum_{k\in[\mathcal{F}]}\hat{\mu}_{k}
\end{equation}
\end{theorem}

In this estimator, the first term ``pretends" that $\hat{\mu}_{\cdot}$ are perfect estimator for the population mean and calculate the variance, while the second term corrects for the fact that the empirical mean estimation is not perfect.
Notice the theorem only requires that $\hat{\mu}$ and $\hat{\phi}$ are unbiased, and is agnostic to the actual computation procedure by these estimators.

\paragraph{Proof} 
We define the population mean of $\mathcal{D}_{k}$ to be $\mu_{k}$, i.e.
\begin{equation}
    \mu_{k} := \mathbb{E}_{X\sim \mathcal{D}_{k}}[X],
\end{equation}
and the population mean of $\mu_{k}$ across randomness in $\mathcal{D}$ to be $\mu$, i.e.
\begin{equation}
    \mu := \mathbb{E}_{\mathcal{D}\sim\Xi}[\mathbb{E}_{X\sim \mathcal{D}}[X]]
\end{equation}

We look at the first term of the estimator in equation \ref{eq:unbiased-variance-across-mean}:

\begin{align} \label{eq:1}
    &\frac{1}{\mathcal{F} - 1}\mathbb{E}[\sum_{k\in [\mathcal{F}]}(\hat{\mu}_{k}-\hat{\mu})^{2}] \\
    &= \frac{1}{\mathcal{F} - 1}\mathbb{E}[\sum_{k\in [\mathcal{F}]}((\hat{\mu}_{k}- \mu_{k}) - (\hat{\mu} - \mu) \nonumber \\
    &+ (\mu_{k} - \mu))^{2}] \nonumber \\
    &= \frac{1}{\mathcal{F} - 1}\mathbb{E}[\sum_{k\in [\mathcal{F}]}[
    (\hat{\mu}_{k}- \mu_{k})^{2} + (\hat{\mu} - \mu))^{2}\nonumber \\
    &+ (\mu_{k} - \mu)^{2} - 2(\hat{\mu}_{k}-\mu_{k})(\hat{\mu} - \mu)\nonumber \\
    &- 2(\hat{\mu_{k}}- \mu_{k})(\hat{\mu} - \mu)]] \nonumber 
\end{align}

There are 5 summands within $\sum_{k\in[\mathcal{F}]}$, and we look at them one by one:

\begin{equation}
    \mathbb{E}[\sum_{k\in [\mathcal{F}]}(\hat{\mu}_{k} - \mu_{k})^{2}] = \mathbb{E}[\sum_{k\in [\mathcal{F}]}\hat{\phi}_{k}],
\end{equation}

\begin{align}\label{eq:unbiased-mean-estimator-variance}
    \mathbb{E}[(\hat{\mu} - \mu)^{2}] &= \mathbb{E}[(\mu - \frac{1}{\mathcal{F}}\sum_{k\in[\mathcal{F}]}\mu_{k}) \\
    & + \frac{1}{\mathcal{F}}\sum_{k\in[\mathcal{F}]}(\mu_{k} - \hat{\mu}_{k}))^{2}] \nonumber \\
    &= \frac{1}{\mathcal{F}}V + \frac{1}{\mathcal{F}^{2}}\sum_{k\in [\mathcal{F}]}\mathbb{E}[\hat{\phi}_{k}] \nonumber 
\end{align}

\begin{equation}
    \mathbb{E}[\sum_{k\in[\mathcal{F}]}(\mu_{k} - \mu)^{2}] = \mathcal{F}V
\end{equation}

\begin{align}
    &\mathbb{E}[-2\sum_{k\in[\mathcal{F}]}(\hat{\mu}_{k} - \mu_{k})(\hat{\mu} - \mu)] \\
    &= -\frac{2}{\mathcal{F}}\mathbb{E}[\sum_{k\in [\mathcal{F}]}\hat{\phi}_{k}] \nonumber .
\end{align}

\begin{align}
    &\mathbb{E}[-2\sum_{k\in [\mathcal{F}]}(\hat{\mu}_{k} - \mu_{k})(\hat{\mu} - \mu)] \nonumber \nonumber \\
    &= -2V.
\end{align}

Putting these five terms together, we continue calculating Equation \ref{eq:1}:
\begin{align}
    &\frac{1}{\mathcal{F} - 1}\mathbb{E}[\sum_{k\in [\mathcal{F}]}(\hat{\mu}_{k}-\hat{\mu})^{2}] \\
    &= \frac{1}{\mathcal{F} - 1}\mathbb{E}[\sum_{k\in [
    \mathcal{F}]}\hat{\phi}_{k} \nonumber\\
    &+ \mathcal{F}(\frac{1}{\mathcal{F}}V
    + \frac{1}{\mathcal{F}^{2}}\sum_{k\in [\mathcal{F}]}\mathbb{E}[\hat{\phi}_{k}]) \nonumber \\
    & + \mathcal{F}V \nonumber\\
    & -\frac{2}{\mathcal{F}} \sum_{k\in [\mathcal{F}]} \hat{\phi}_{k} \nonumber\\
    & - 2V
    ] \nonumber\\
    &= V + \mathbb{E}[\frac{1}{\mathcal{F}} \sum_{k}\hat{\phi_{k}}]\nonumber
\end{align}

Then from Equation \ref{eq:unbiased-variance-across-mean}, we can tell that $\hat{Var}_{F}$ is unbiased. $\square$

Now we come back to the topic of developing an unbiased estimator for $Var_{F}$ as defined in Equation \ref{eq:varf}. To utilize Theorem \ref{bias-variance-thm-1}, we need two components:
\begin{itemize}
    \item An unbiased estimator $\hat{\mu}_{F_{k}}$ for $\mathbb{E}_{E}[c_{F,E}|F=F_{k}]$
    \item An unbiased estimator $\hat{\phi}_{F_{k}}$ for the variance of $\hat{\mu}_{F_{k}}$, i.e. $Var_{E_{(\cdot})}(\hat{\mu}_{F_{k}})$
\end{itemize}

$\Bar{c}_{F_{k}}$ is an unbiased estimator for $\mathbb{E}_{E}[c_{F,E}|F=F_{k}]$, and its variance $Var_{E}[\Bar{c}_{F}|F=F_{k}]$ is
\begin{equation}
    Var_{E_{(\cdot)}}(\Bar{c}_{F_{k}}) = \frac{1}{\mathcal{E}}Var_{E}[c_{F,E}|F=F_{k}].
\end{equation}

Therefore, to develop an unbiased estimator for $Var_{E}(\Bar{c}_{F_{jk}})$, it suffices to have an unbiased estimate of $Var_{E}[c_{F,E}|F=F_{k}]$.
We define
\begin{equation} \label{eq:simplest-variance-estimate}
    \hat{\phi}_{F_{k}} := \frac{1}{\mathcal{E}(\mathcal{F} - 1)}\sum_{k\in[\mathcal{L}]}(c_{F_{k},E_{kl}} - \Bar{c}_{F_{k}})^{2},
\end{equation}

and we can plug in $\hat{\phi}_{F_{k}}$ and $\hat{\mu}_{F_{k}} = \Bar{c}_{F_{k}}$ into Theorem \ref{bias-variance-thm-1} as an unbiased estimator to obtain an unbiased estimator for $Var_{F}[\mathbb{E}_{E}[c_{P,F,E}]|P=P_{j}]$, and we average the estimation for each $P_{j}$ to obtain an unbiased estimate.

\subsubsection{Estimating $\pretrainvar$} \label{sec:estimatingpretrainvar}
We next estimate $Var_{P}[\mathbb{E}_{F,E}[c_{P,F,E}]]$
We can still apply the idea from Theorem \ref{bias-variance-thm-1}, which requires
\begin{itemize}
    \item An unbiased estimator $\hat{\mu}_{P_{j}}$ for $\mathbb{E}_{F,E}[c_{P,F,E}|P=P_{j}]$
    \item An unbiased estimator $\hat{\phi}_{P_{j}}$ for the variance of $\hat{\mu}_{P_{j}}$, i.e. $Var_{F,E}[\hat{\mu}_{P_{j}}]$.
\end{itemize}
Again, the first is easy to obtain: $\hat{\mu}_{P_{j}} = \Bar{c}_{P_{j}}$ is an unbiased estimator for $\mathbb{E}_{F,E}[c_{P,F,E}|P=P_{j}]$, where
\begin{equation}
    \Bar{c}_{P_{j}} := \frac{1}{\mathcal{F}\mathcal{E}}\sum_{k\in [\mathcal{F}], l \in [\mathcal{E}]}c_{P_{j},F_{jk},E_{jkl}}
\end{equation}

However, we cannot straightforwardly estimate $Var_{F,E}[\hat{\mu}_{P_{j}}]$ as before, since samples $c_{P_{j}, F_{jk}, E_{jkl}}$ are no longer independent. 
We need to use Equation \ref{eq:unbiased-mean-estimator-variance} to develop an unbiased estimator (the LHS is exactly what we want!), i.e.
\begin{align} \label{eq:example-recurse}
    Var_{F,E}(\Bar{c}_{P_{j}}) &= \frac{1}{\mathcal{F}}Var_{F}(\mathbb{E}_{E}[c_{P,F,E}]|P=P_{j}) \\
    &+ \frac{1}{\mathcal{F}^{2}}\sum_{k\in [\mathcal{F}]}Var_{E}(\Bar{c}_{P_{j},F_{jk}})\nonumber,
\end{align}
and we already know how to estimate these two summands from the previous discussion on estimating $\finetunevar$.

\subsubsection{Estimating $\biassquared$}
It is easy to see that the following $\hat{L}$ is an unbiased estimator for the loss $\mathcal{L}$.

\begin{equation}
    \hat{\mathcal{L}} := \frac{1}{\mathcal{P}\mathcal{F}\mathcal{E}}\sum_{j\in[\mathcal{P}], k\in[\mathcal{F}], l\in[\mathcal{E}]}(1 - c_{P_{j},F_{jk},E_{jkl}})^{2},
\end{equation}

and 
\begin{equation}
    \mathbb{E}[\hat{\mathcal{L}}] = \mathcal{L}.
\end{equation}

By linearity of expectation and loss decomposition in Equation \ref{eq:decomposition}, 
\begin{align}
    \hat{\biassquared} &:= \hat{\mathcal{L}} - \hat{\pretrainvar} \\
    &- \hat{\finetunevar} - \hat{\ckptvar} \nonumber
\end{align}
is an unbiased estimator of $\biassquared$.

Notice that the naïve estimator that calculates the expected bias and then squares it estimates $(\mathbb{E}[\bias])^{2}$ instead of $\mathbb{E}[\biassquared]$.

\subsection{Generalization} \label{sec:decompoe-general}
We can generalize this estimation strategy to decompose variance into arbitrary number of randomness.
In general, we want to estimate some quantity of the following form
\begin{equation}
    \mathbb{E}_{r_{1} \dots, r_{n-1}}[Var_{r_{n}}[\mathbb{E}_{r_{n+1}\dots r_{N}}[c_{r_{1}\dots, c_{N}}]]],
\end{equation}

from the data that has an hierarchical tree structure of randomness. 

For the goal of developing an unbiased estimator, we can get rid of the outer expectation $r_{1} \dots r_{n-1}$ easily by linearity of expectation: simply estimate the Variance conditioned on $r_{1\dots n-1}$ and average them together, as discussed in Section \ref{sec:estimatingckpttunevar}.

To estimate 
\begin{equation}
    Var_{r_{n}}[\mathbb{E}_{r_{n+1}\dots r_{N}}[c_{r_{1}\dots, c_{N}}]],
\end{equation}
we make use of Theorem \ref{bias-variance-thm-1}, which requires
\begin{itemize}
    \item an unbiased estimator $\hat{\mu}_{r_{n+1}}$ for the quantity $\mathbb{E}_{r_{n+1}\dots r_{N}}[c_{r_{1}\dots, c_{N}}]$, which we can straightforwardly obtain by average the examples that has the same random variables $r_{1\dots n}$ (e.g. $\Bar{c}_{P_{j}}$) 
    \item an unbiased estimator for the variance of $\hat{\mu}_{r_{n+1}}$. If $N = n + 1$, we can directly compute the sample variance of the $c$ as our estimate (e.g. in Equation \ref{eq:simplest-variance-estimate}). Otherwise, we use Equation \ref{eq:unbiased-mean-estimator-variance} to decompose the desired quantities into two, and estimate them recursively by applying Theorem \ref{bias-variance-thm-1} and Equation \ref{eq:unbiased-mean-estimator-variance}.
\end{itemize}

For readability we wrote the proof with the assumption that, in the tree of randomness, the number of branches for each node at the same depth is the same. 
However, our proof does not make use of this assumption and can be applied to a general tree structure of randomness as long as the the number of children is larger or equal to 2 for each non-terminal node.

\section{Variance Conditioned on Bias} \label{sec:var-conditioned-on-bias}
Since lower bias usually implies lower variance, to tease out the latent effect, we estimate the variance ``given a fixed level of bias $\biassquared$ of $b^{2} \in [0, 1]$", i.e.

\begin{align}
    \pretrainvar(b^{2}) &:= \mathbb{E}_{i}[\pretrainvar_{i}|\biassquared_{i}=b^2]
\end{align}

\begin{figure}%
    \centering
    \subfloat[\centering]{{\includegraphics[width=3.5cm]{Instance-level/figs/instance_diff_appendix/mnli_pretrain_var.png}}}%
    \quad
    \subfloat[\centering]{{\includegraphics[width=3.5cm]{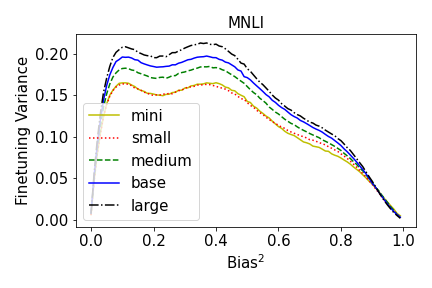}}}%
    \quad
    \subfloat[\centering]{{\includegraphics[width=3.5cm]{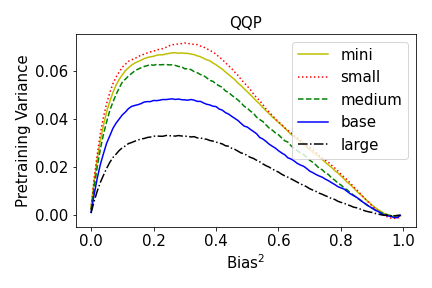}}}%
    \quad
    \subfloat[\centering]{{\includegraphics[width=3.5cm]{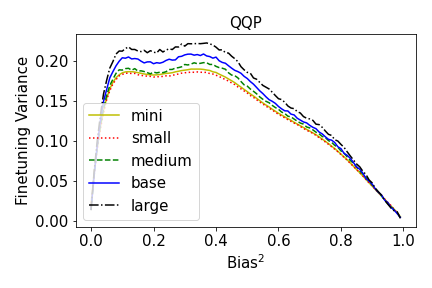}}}%
    \quad
    \subfloat[\centering]{{\includegraphics[width=3.5cm]{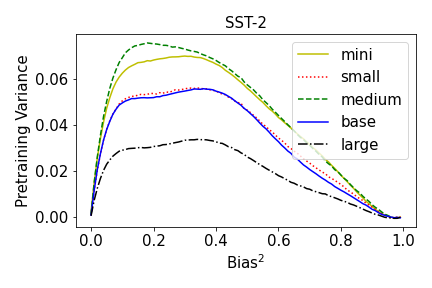}}}%
    \quad
    \subfloat[\centering]{{\includegraphics[width=3.5cm]{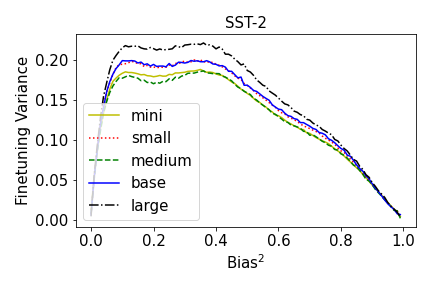}}}%
    \quad
    \caption{The variance curve conditioned on $\biassquared$ for in-domain development set of MNLI, QQP and SST-2. Each curve represents a model size. Left for pretraining variance and right for finetuning variance.}%
    \label{fig:bias-var-curve}%
\end{figure}

We estimate $\pretrainvar^{s}(b^{2})$ and $\finetunevar^{s}(b^{2})$ using gaussian process and plot them against $b^{2}$ in Figure \ref{fig:bias-var-curve} for MNLI, QQP, and SST-2. 
We find that larger models usually have larger finetuning variance across all levels of biases (except for \mediumsize and \minisize on SST-2), and \basesize model always has larger pretraining variance than \largesize. 

\begin{figure}%
    \centering
    \subfloat[\centering]{{\includegraphics[width=3.5cm]{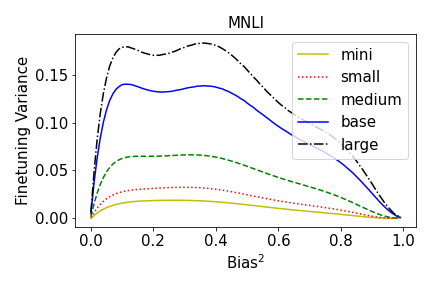}}}%
    \quad
    \subfloat[\centering]{{\includegraphics[width=3.5cm]{Instance-level/figs/instance_diff_appendix/prob_mnli_finetune_var.png}}}%
    \quad
    \subfloat[\centering]{{\includegraphics[width=3.5cm]{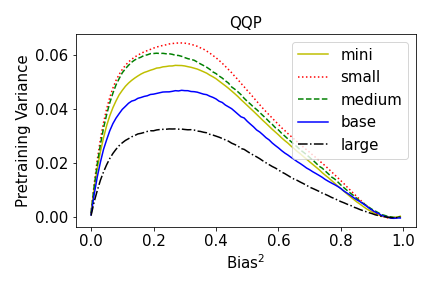}}}%
    \quad
    \subfloat[\centering]{{\includegraphics[width=3.5cm]{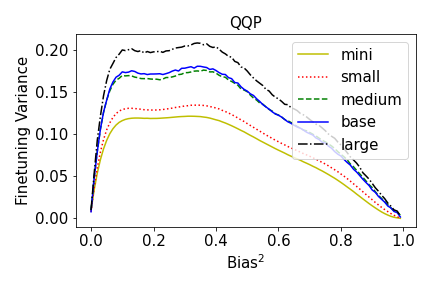}}}%
    \quad
    \subfloat[\centering]{{\includegraphics[width=3.5cm]{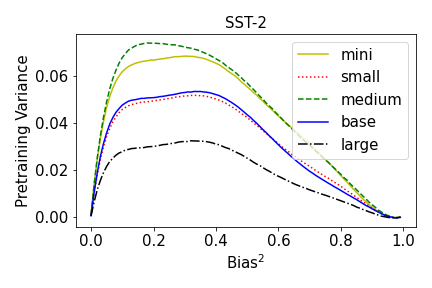}}}%
    \quad
    \subfloat[\centering]{{\includegraphics[width=3.5cm]{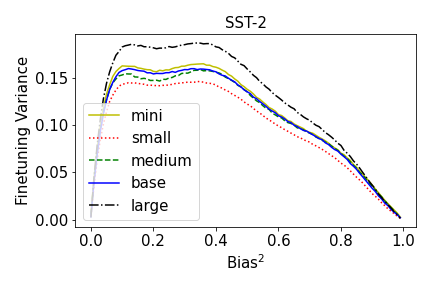}}}%
    \quad
    \caption{The same figure as \ref{fig:bias-var-curve}, except for using the squared loss function $\mathcal{L} = (1-p)^{2}$, where $p$ is the probability assigned to the correct label, instead of 0/1 loss.}%
    \label{fig:prob-bias-var-curve}%
\end{figure}

We also experimented with the squared loss:
\begin{equation}
    \mathcal{L}_{i} = (1 - p_{i})^{2},
\end{equation}
where $p_{i}$ is the probability the assigned to the correct label for instance $i$. 
We plot the same curve in Figure \ref{fig:prob-bias-var-curve} and observe the same trend.

\section{Example Decaying Instances} \label{sec:decaying-instances-all}
We manually examined the group of instances where  $^{\minisize}_{\largesize}\hat{\Delta\acc}_{i} \leq -0.9$ in Table \ref{tab:minivs.large}.
In other words, \minisize is almost always correct on these instances, while \largesize is almost always wrong. 
For each instance in this group, we manually categorize it into one of the four categories: 1) Correct, if the label is correct, 2) Fine, if the label might be controversial but we could see a reason why this label is reasonable, 3) Wrong, if the label is wrong, and 4) Unsure, if we are unsure how to label this instance.
As a control, we also examined the remaining fraction of the dataset.
Each time we annotate an instance, with $50\%$ probability it is sampled  from the decaying fraction or the remaining fraction, and we do not know which group it comes from. 

We show below all the annotated instances from this decaying fraction and their categories for MNLI (Section \ref{sec:decaying-mnli}), QQP , and SST-2(Section \ref{sec:decaying-sst-2}).

\subsection{MNLI} \label{sec:decaying-mnli}
MNLI is the abbreviation of Multi-Genre Natural Language Inference (\citet{williams-etal-2020-predicting}). 
In this task, given a premise and a hypothesis, the model needs to classify whether the premise entails/contradicts the hypothesis, or otherwise.
The instances can be seen below.
\\\\
\textbf{Premise} : and that you're very much right but the jury may or may not see it that way so you get a little anticipate you know anxious there and go well you know \\
\textbf{Hypothesis}  : Jury's operate without the benefit of an education in law. \\
\textbf{Label}  : Neutral \\
\textbf{Category}  : {\color{blue}Correct} \\
\\
\textbf{Premise} : In fiscal year 2000, it reported estimated improper Medicare Fee-for-Service payments of \$11. \\
\textbf{Hypothesis}  : The payments were improper. \\
\textbf{Label}  : Entailment \\
\textbf{Category}  : Fine \\
\\
\textbf{Premise} : is that what you ended up going into \\
\textbf{Hypothesis}  : So that must be what you chose to do? \\
\textbf{Label}  : Entailment \\
\textbf{Category}  : {\color{blue}Correct} \\
\\
\textbf{Premise} : INTEREST RATE - The price charged per unit of money borrowed per year, or other unit of time, usually expressed as a percentage. \\
\textbf{Hypothesis}  : Interest rate is defined as the total amount of money borrowed.  \\
\textbf{Label}  : Entailment \\
\textbf{Category}  : {\color{red}Wrong} \\
\\
\textbf{Premise} : The analyses comply with the informational requirements of the sections including the classes of small entities subject to the rule and alternatives considered to reduce the burden on the small entities. \\
\textbf{Hypothesis}  : The rules place a high burden on the activities of small entities. \\
\textbf{Label}  : Contradiction \\
\textbf{Category}  : {\color{blue}Correct} \\
\\
\textbf{Premise} : Isn't a woman's body her most personal property? \\
\textbf{Hypothesis}  : Women's bodies belong to themselves, they should decide what to do with it.  \\
\textbf{Label}  : Neutral \\
\textbf{Category}  : Unsure \\
\\
\textbf{Premise} : The Standard , published a few days before Deng's death, covers similar territory. \\
\textbf{Hypothesis}  : The Washington Post covers similar territory. \\
\textbf{Label}  : Neutral \\
\textbf{Category}  : {\color{blue}Correct} \\
\\
\textbf{Premise} : Shoot only the ones that face us, Jon had told Adrin. \\
\textbf{Hypothesis}  : Jon told Adrin and the others to only shoot the ones that face us. \\
\textbf{Label}  : Entailment \\
\textbf{Category}  : {\color{red}Wrong} \\
\\
\textbf{Premise} : But if you take it seriously, the anti-abortion position is definitive by definition. \\
\textbf{Hypothesis}  : If you decide to be serious about supporting anti-abortion, it's a very run of the mill belief to hold. \\
\textbf{Label}  : Neutral \\
\textbf{Category}  : Unsure \\
\\
\textbf{Premise} : yeah well that's the other thing you know they talk about women leaving the home and going out to work well still taking care of the children is a very important job and and someone's got to do it and be able to do it right and \\
\textbf{Hypothesis}  : It is not acceptable for anybody to refuse work in order to take care of children. \\
\textbf{Label}  : Contradiction \\
\textbf{Category}  : {\color{blue}Correct} \\
\\
\textbf{Premise} : The researchers found expected stresses like the loss of a check in the mail and the illness of loved ones. \\
\textbf{Hypothesis}  : The stresses affected people much diffferently than the researchers expected. \\
\textbf{Label}  : Contradiction \\
\textbf{Category}  : {\color{blue}Correct} \\
\\
\textbf{Premise} : so you know it's something we we have tried to help but yeah \\
\textbf{Hypothesis}  : We did what we could to help. \\
\textbf{Label}  : Entailment \\
\textbf{Category}  : {\color{blue}Correct} \\
\\
\textbf{Premise} : Czarek was welcomed enthusiastically, even though the poultry brotherhood was paying a lot of sudden attention to the newcomers - a strong group of young and talented managers from an egzemo-exotic chicken farm in Fodder Band nearby Podunkowice. \\
\textbf{Hypothesis}  : Czarek was welcomed into the group by the farmers. \\
\textbf{Label}  : Entailment \\
\textbf{Category}  : {\color{blue}Correct} \\
\\
\textbf{Premise} : 'I don't suppose you could forget I ever said that?' \\
\textbf{Hypothesis}  : I hope that you can remember that forever.  \\
\textbf{Label}  : Contradiction \\
\textbf{Category}  : {\color{red}Wrong} \\
\\
\textbf{Premise} : Oh, my friend, have I not said to you all along that I have no proofs.  \\
\textbf{Hypothesis}  : I told you from the start that I had no evidence. \\
\textbf{Label}  : Entailment \\
\textbf{Category}  : {\color{blue}Correct} \\
\\
\textbf{Premise} : I should put it this way.  \\
\textbf{Hypothesis}  : I should phrase it differently. \\
\textbf{Label}  : Entailment \\
\textbf{Category}  : {\color{blue}Correct} \\
\\
\textbf{Premise} : An organization's activities, core processes, and resources must be aligned to support its mission and help it achieve its goals. \\
\textbf{Hypothesis}  : An organization is successful if its activities, resources, and goals align. \\
\textbf{Label}  : Entailment \\
\textbf{Category}  : Fine \\
\\
\textbf{Premise} : A more unusual dish is azure, a kind of sweet porridge made with cereals, nuts, and fruit sprinkled with rosewater. \\
\textbf{Hypothesis}  : Azure is a common and delicious food made with cereals, nuts and fruit. \\
\textbf{Label}  : Entailment \\
\textbf{Category}  : {\color{red}Wrong} \\
\\
\textbf{Premise} : once you have something and it's like i was watching this program on TV yesterday in nineteen seventy six NASA came up with Three D graphics right \\
\textbf{Hypothesis}  : I was watching a program about gardening.  \\
\textbf{Label}  : Contradiction \\
\textbf{Category}  : {\color{blue}Correct} \\
\\
\textbf{Premise} : , First-Class Mail used by households to pay their bills) and the household bill mail (i.e. \\
\textbf{Hypothesis}  : Second-Class Mail used by households to pay their bills \\
\textbf{Label}  : Contradiction \\
\textbf{Category}  : Unsure \\
\\
\textbf{Premise} : Rightly or wrongly, America is seen as globalization's prime mover and head cheerleader and will be blamed for its excesses until we start paying official attention to them. \\
\textbf{Hypothesis}  : America's role in the globalization movement is important whether we agree with it or not.  \\
\textbf{Label}  : Entailment \\
\textbf{Category}  : {\color{blue}Correct} \\
\\
\textbf{Premise} : After being diagnosed with cancer, Carrey's Kaufman decides to do a show at Carnegie Hall. \\
\textbf{Hypothesis}  : Carrey's Kaufman is only diagnosed with cancer after doing a show at Carnegie Hall. \\
\textbf{Label}  : Contradiction \\
\textbf{Category}  : {\color{blue}Correct} \\
\\
\textbf{Premise} : Several pro-life Dems are mounting serious campaigns at the state level, often against pro-choice Republicans. \\
\textbf{Hypothesis}  : Serious campaigns are being run by a few pro-life Democrats. \\
\textbf{Label}  : Entailment \\
\textbf{Category}  : {\color{blue}Correct} \\
\\
\textbf{Premise} : On the northwestern Alpine frontier, a new state had appeared on the scene, destined to lead the movement to a united Italy. \\
\textbf{Hypothesis}  : The unite Italy movement was waiting for a leader.  \\
\textbf{Label}  : Neutral \\
\textbf{Category}  : Fine \\
\\
\textbf{Premise} : well we bought this with credit too  well we found it with a clearance uh down in Memphis i guess and uh \\
\textbf{Hypothesis}  : We bought non-sale items in Memphis on credit. \\
\textbf{Label}  : Contradiction \\
\textbf{Category}  : {\color{blue}Correct} \\
\\
\textbf{Premise} : He slowed. \\
\textbf{Hypothesis}  : He stopped moving so quickly. \\
\textbf{Label}  : Entailment \\
\textbf{Category}  : {\color{blue}Correct} \\
\\
\textbf{Premise} : As legal scholar Randall Kennedy wrote in his book Race, Crime, and the Law , Even if race is only one of several factors behind a decision, tolerating it at all means tolerating it as potentially the decisive factor. \\
\textbf{Hypothesis}  : Race is one of several factors in some judicial decisions \\
\textbf{Label}  : Entailment \\
\textbf{Category}  : {\color{blue}Correct} \\
\\
\textbf{Premise} : Although all four categories of emissions are down substantially, they only achieve 50-75\% of the proposed cap by 2007 (shown as the dotted horizontal line in each of the above figures). \\
\textbf{Hypothesis}  : All of the emission categories experienced a downturn except for one. \\
\textbf{Label}  : Contradiction \\
\textbf{Category}  : {\color{blue}Correct} \\
\\
\textbf{Premise} : He sat up, trying to free himself. \\
\textbf{Hypothesis}  : He was trying to take a nap. \\
\textbf{Label}  : Contradiction \\
\textbf{Category}  : {\color{blue}Correct} \\
\\
\textbf{Premise} : Impossible. \\
\textbf{Hypothesis}  : Cannot be done. \\
\textbf{Label}  : Entailment \\
\textbf{Category}  : {\color{blue}Correct} \\
\\
\textbf{Premise} : But, as the last problem I'll outline suggests, neither of the previous two objections matters. \\
\textbf{Hypothesis}  : I will not continue to outline any more problems. \\
\textbf{Label}  : Entailment \\
\textbf{Category}  : {\color{blue}Correct} \\
\\
\textbf{Premise} : As the Tokugawa shoguns had feared, this opening of the floodgates of Western culture after such prolonged isolation had a traumatic effect on Japanese society. \\
\textbf{Hypothesis}  : The Tokugawa shoguns had feared that, because they understood the Japanese society very well. \\
\textbf{Label}  : Neutral \\
\textbf{Category}  : Fine \\
\\
\textbf{Premise} : In the ancestral environment a man would be likely to have more offspring if he got his pick of the most fertile-seeming women. \\
\textbf{Hypothesis}  : Only a man who stayed with one female spread his genes most efficiently. \\
\textbf{Label}  : Contradiction \\
\textbf{Category}  : Fine \\
\\
\textbf{Premise} : Tommy was suddenly galvanized into life. \\
\textbf{Hypothesis}  : Tommy had been downcast for days. \\
\textbf{Label}  : Neutral \\
\textbf{Category}  : {\color{blue}Correct} \\
\\
\textbf{Premise} : Improved products and services   Initiate actions and manage risks to develop new products and services within or outside the organization. \\
\textbf{Hypothesis}  : Managed risks lead to new products \\
\textbf{Label}  : Entailment \\
\textbf{Category}  : Fine \\
\\
\textbf{Premise} : Coast Guard rules establishing bridgeopening schedules). \\
\textbf{Hypothesis}  : The Coast Guard is in charge of opening bridges. \\
\textbf{Label}  : Entailment \\
\textbf{Category}  : {\color{blue}Correct} \\
\\
\textbf{Premise} : The anthropologist Napoleon Chagnon has shown that Yanomamo men who have killed other men have more wives and more offspring than average guys. \\
\textbf{Hypothesis}  : Yanomamo men who kill other men have better chances at getting more wives. \\
\textbf{Label}  : Entailment \\
\textbf{Category}  : Fine \\
\\
\textbf{Premise} : The Varanasi Hindu University has an Art Museum with a superb collection of 16th-century Mughal miniatures, considered superior to the national collection in Delhi. \\
\textbf{Hypothesis}  : The Varanasi Hindu University has an art museum on its campus which may be superior objectively to the national collection in Delhi. \\
\textbf{Label}  : Entailment \\
\textbf{Category}  : {\color{blue}Correct} \\
\\
\textbf{Premise} : Part of the reason for the difference in pieces per possible delivery may be due to the fact that five percent of possible residential deliveries are businesses, and it is thought, but not known, that a lesser percentage of possible deliveries on rural routes are businesses. \\
\textbf{Hypothesis}  : We all know that the reason for a lesser percentage of possible deliveries on rural routes being businesses, is because of the fact that people prefer living in cities rather than rural areas. \\
\textbf{Label}  : Neutral \\
\textbf{Category}  : {\color{blue}Correct} \\
\\
\textbf{Premise} : right oh they've really done uh good job of keeping everybody informed of what's going on sometimes i've wondered if it wasn't almost more than we needed to know \\
\textbf{Hypothesis}  : I don't think I have shared enough information with everyone.  \\
\textbf{Label}  : Contradiction \\
\textbf{Category}  : {\color{blue}Correct} \\
\\
\textbf{Premise} : To reach any of the three Carbet falls, you must continue walking after the roads come to an end for 20 minutes, 30 minutes, or two hours respectively. \\
\textbf{Hypothesis}  : There are three routes to the three Carbet falls, each a different length and all continue after the road seemingly ends. \\
\textbf{Label}  : Entailment \\
\textbf{Category}  : {\color{blue}Correct} \\
\\
\textbf{Premise} : But when the cushion is spent in a year or two, or when the next recession arrives, the disintermediating voters will find themselves playing the roles of budget analysts and tax wonks. \\
\textbf{Hypothesis}  : The cushion will likely be spent in under two years. \\
\textbf{Label}  : Entailment \\
\textbf{Category}  : {\color{blue}Correct} \\
\\
\textbf{Premise} : But, Slate protests, it was [Gates'] byline that appeared on the cover. \\
\textbf{Hypothesis}  : Slate was one hundred percent positive it was Gates' byline on the cover. \\
\textbf{Label}  : Neutral \\
\textbf{Category}  : {\color{blue}Correct} \\
\\
\textbf{Premise} : But it's for us to get busy and do something." \\
\textbf{Hypothesis}  : "We don't do much, so maybe this would be good for us to bond and be together for the first time in a while.". \\
\textbf{Label}  : Neutral \\
\textbf{Category}  : Fine \\
\\
\textbf{Premise} : Pearl Jam detractors still can't stand singer Eddie  They say he's unbearably self-important and limits the group's appeal by refusing to sell out and make videos. \\
\textbf{Hypothesis}  : A lot of people consider Eddie to be a bad singer. \\
\textbf{Label}  : Neutral \\
\textbf{Category}  : {\color{blue}Correct} \\
\\
\textbf{Premise} : it's the very same type of paint and everything \\
\textbf{Hypothesis}  : It's the same paint formula, it's great! \\
\textbf{Label}  : Entailment \\
\textbf{Category}  : Fine \\
\\
\textbf{Premise} : Exhibit 3 presents total national emissions of NOx and SO2 from all sectors, including power. \\
\textbf{Hypothesis}  : In Exhibit 3 there are the total regional emissions od NOx and SO2 from all sectors. \\
\textbf{Label}  : Entailment \\
\textbf{Category}  : {\color{blue}Correct} \\
\\
\textbf{Premise} : uh-huh and is it true i mean is it um \\
\textbf{Hypothesis}  : It's true. \\
\textbf{Label}  : Entailment \\
\textbf{Category}  : {\color{red}Wrong} \\
\\
\textbf{Premise} : When a GAGAS attestation engagement is the basis for an auditor's subsequent report under the AICPA standards, it would be advantageous to users of the subsequent report for the auditor's report to include the information on compliance with laws and regulations and internal control that is required by GAGAS but not required by AICPA standards. \\
\textbf{Hypothesis}  : The report is required by GAGAS but not AICPA. \\
\textbf{Label}  : Entailment \\
\textbf{Category}  : {\color{blue}Correct} \\
\\
\textbf{Premise} : i'm on i'm in the Plano school system and living in Richardson and there is a real dichotomy in terms of educational and economic background of the kids that are going to be attending this school \\
\textbf{Hypothesis}  : The Plano school system only has children with poor intelligence. \\
\textbf{Label}  : Contradiction \\
\textbf{Category}  : {\color{blue}Correct} \\
\\

\subsection{QQP} \label{sec:decaying-qqp}
QQP is the abbreviation of Quora Question Pairs\footnote{https://www.quora.com/q/quoradata/First-Quora-Dataset-Release-Question-Pairs}. 
Given two questions, the model needs to tell whether they have the same meaning (i.e. Paraphrase/Non-paraphrase).
\\
\textbf{Question 1} : Which universities for MS in CS should I apply to? \\
\textbf{Question 2} : Which universities should I apply to for an MS in CS? \\
\textbf{Label} : Paraphrase \\
\textbf{Category} : {\color{blue}Correct} \\
\\
\textbf{Question 1} : What should I do to make life worth living? \\
\textbf{Question 2} : What makes life worth living? \\
\textbf{Label} : Paraphrase \\
\textbf{Category} : Fine \\
\\
\textbf{Question 1} : Why did Quora remove my question? \\
\textbf{Question 2} : Why does Quora remove questions? \\
\textbf{Label} : Paraphrase \\
\textbf{Category} : {\color{blue}Correct} \\
\\
\textbf{Question 1} : How do I get thousands of followers on Instagram? \\
\textbf{Question 2} : How can I get free 10k real Instagram followers fast? \\
\textbf{Label} : Paraphrase \\
\textbf{Category} : Fine \\
\\
\textbf{Question 1} : What is the basic knowledge of computer science engineers? \\
\textbf{Question 2} : What is basic syallbus of computer science engineering? \\
\textbf{Label} : Non-paraphrase \\
\textbf{Category} : Fine \\
\\
\textbf{Question 1} : How many mosquito bites does it take to kill a human being? \\
\textbf{Question 2} : How many times can a single mosquito bite a human within 8 hours? \\
\textbf{Label} : Non-paraphrase \\
\textbf{Category} : {\color{blue}Correct} \\
\\
\textbf{Question 1} : How does it feel to become attractive from unattractive? \\
\textbf{Question 2} : What does it feel like to go from physically unattractive to physically attractive? \\
\textbf{Label} : Paraphrase \\
\textbf{Category} : {\color{blue}Correct} \\
\\
\textbf{Question 1} : Who is answering the questions asked on Quora? \\
\textbf{Question 2} : Who can answer the questions asked on Quora? \\
\textbf{Label} : Paraphrase \\
\textbf{Category} : {\color{blue}Correct} \\
\\
\textbf{Question 1} : What machine learning theory do I need to know in order to be a successful machine learning practitioner? \\
\textbf{Question 2} : What do I need to know to learn machine learning? \\
\textbf{Label} : Paraphrase \\
\textbf{Category} : {\color{blue}Wrong} \\
\\
\textbf{Question 1} : If you could go back in time and change one thing, what would it be and why? \\
\textbf{Question 2} : If you could go back in time and do one thing, what would it be? \\
\textbf{Label} : Paraphrase \\
\textbf{Category} : {\color{blue}Correct} \\
\\
\textbf{Question 1} : Will there be a civil war if Trump doesn't become president? \\
\textbf{Question 2} : Will there be a second civil war if Trump becomes president? \\
\textbf{Label} : Paraphrase \\
\textbf{Category} : {\color{blue}Correct} \\
\\
\textbf{Question 1} : Do Quora contributors get paid? \\
\textbf{Question 2} : How do contributors get paid by Quora? \\
\textbf{Label} : Paraphrase \\
\textbf{Category} : {\color{blue}Correct} \\
\\
\textbf{Question 1} : Did India meet Abdul Kalam's 2020 vision so far? \\
\textbf{Question 2} : How far do you think India has reached on President APJ Kalam's vision in the book India 2020? \\
\textbf{Label} : Non-paraphrase \\
\textbf{Category} : {\color{blue}Correct} \\
\\
\textbf{Question 1} : How do I stop my dog from whining after getting spayed? \\
\textbf{Question 2} : How do I stop my dog from whining? \\
\textbf{Label} : Paraphrase \\
\textbf{Category} : {\color{blue}Wrong} \\
\\
\textbf{Question 1} : What difference are exactly between Euclidean space and non Euclidean space? \\
\textbf{Question 2} : What is the difference between Euclidean and non-Euclidean? \\
\textbf{Label} : Non-paraphrase \\
\textbf{Category} : {\color{blue}Wrong} \\
\\
\textbf{Question 1} : Why doesn't Hillary Clinton win the White House if she won the popular vote? \\
\textbf{Question 2} : How did Hillary Clinton win the popular vote but Donald Trump win the election? \\
\textbf{Label} : Paraphrase \\
\textbf{Category} : {\color{blue}Correct} \\
\\
\textbf{Question 1} : How is public breastfeeding seen where you live? \\
\textbf{Question 2} : How is breastfeeding in public seen in your country? \\
\textbf{Label} : Paraphrase \\
\textbf{Category} : {\color{blue}Correct} \\
\\
\textbf{Question 1} : What are some ways to change your Netflix password? \\
\textbf{Question 2} : How do you change your Netflix password and email? \\
\textbf{Label} : Paraphrase \\
\textbf{Category} : Fine \\
\\
\textbf{Question 1} : What do you think, is your best answer on Quora? \\
\textbf{Question 2} : What is your best answer on Quora? \\
\textbf{Label} : Paraphrase \\
\textbf{Category} : {\color{blue}Correct} \\
\\
\textbf{Question 1} : How can I travel to Mexico without a passport? \\
\textbf{Question 2} : Can I travel to Mexico without a passport? \\
\textbf{Label} : Paraphrase \\
\textbf{Category} : {\color{blue}Correct} \\
\\
\textbf{Question 1} : How do modern Congolese people view Mobutu in retrospect? \\
\textbf{Question 2} : How do Congolese currently view Mobutu Sese Seko? \\
\textbf{Label} : Non-paraphrase \\
\textbf{Category} : {\color{blue}Correct} \\
\\
\textbf{Question 1} : How is Tanmay Bhat losing weight? \\
\textbf{Question 2} : Tanmay Bhat: How did you manage to reduce your fat? \\
\textbf{Label} : Non-paraphrase \\
\textbf{Category} : Unsure \\
\\
\textbf{Question 1} : Is Xiaomi a brand to trust (comparing it with brands like Samsung and HTC)? What is better: Xiaomi MI3 or HTC Desire 816? \\
\textbf{Question 2} : Is xiaomi a trusted brand? \\
\textbf{Label} : Non-paraphrase \\
\textbf{Category} : {\color{blue}Correct} \\
\\
\textbf{Question 1} : Why did Buddhism spread in East Asia and not in its native land India? \\
\textbf{Question 2} : How was Buddhism spread in Asia? \\
\textbf{Label} : Non-paraphrase \\
\textbf{Category} : {\color{blue}Correct} \\
\\
\textbf{Question 1} : Can I become a multi billionaire betting on horses? \\
\textbf{Question 2} : How much money can I make betting on horses? A month? Can I make 20,000 a month? \\
\textbf{Label} : Paraphrase \\
\textbf{Category} : Fine \\
\\
\textbf{Question 1} : What is a diet for gaining weight? \\
\textbf{Question 2} : What is a way to gain weight? \\
\textbf{Label} : Non-paraphrase \\
\textbf{Category} : {\color{blue}Correct} \\
\\
\textbf{Question 1} : How do I use Instagram on my computer? \\
\textbf{Question 2} : How can I get Instagram on my computer? \\
\textbf{Label} : Paraphrase \\
\textbf{Category} : Fine \\
\\
\textbf{Question 1} : What is the legal basis of a "you break it, you buy it" policy? \\
\textbf{Question 2} : Is a "you break it you buy it" policy actually legal? \\
\textbf{Label} : Paraphrase \\
\textbf{Category} : {\color{blue}Correct} \\
\\
\textbf{Question 1} : How I should fix my computer while it is showing no boot device found? \\
\textbf{Question 2} : How do I fix the "Boot device not found" problem? \\
\textbf{Label} : Paraphrase \\
\textbf{Category} : {\color{blue}Correct} \\
\\
\textbf{Question 1} : What innovative name can I use for an interior designing firm? \\
\textbf{Question 2} : What can i name my interior designing firm? \\
\textbf{Label} : Paraphrase \\
\textbf{Category} : {\color{blue}Correct} \\
\\
\textbf{Question 1} : What would it realistically cost to go to Tomorrowland? \\
\textbf{Question 2} : How much is a ticket to Tomorrowland? \\
\textbf{Label} : Non-paraphrase \\
\textbf{Category} : Fine \\
\\
\textbf{Question 1} : Is there a gender pay gap? If so why? \\
\textbf{Question 2} : Is the gender pay gap a myth? \\
\textbf{Label} : Paraphrase \\
\textbf{Category} : {\color{blue}Correct} \\
\\
\textbf{Question 1} : How can I get rid of a canker sore on the bottom of my tongue? \\
\textbf{Question 2} : How can I get rid or a canker sore on the tip of my tongue? \\
\textbf{Label} : Paraphrase \\
\textbf{Category} : {\color{blue}Correct} \\
\\
\textbf{Question 1} : How can I sleep better and early in night? \\
\textbf{Question 2} : How can I sleep better at night? \\
\textbf{Label} : Paraphrase \\
\textbf{Category} : Fine \\
\\
\textbf{Question 1} : Why did DC Change Captain Marvel's name? \\
\textbf{Question 2} : Why did DC have to change Captain Marvel's name but Marvel didn't have to change Scarecrow's name? \\
\textbf{Label} : Paraphrase \\
\textbf{Category} : Fine \\
\\
\textbf{Question 1} : Should be there any difference between IIT and non IIT students in terms of placement package from a company if both of them are equally talented? \\
\textbf{Question 2} : Should be there any difference between IIT and non IIT students in terms of placement package from a company if both of them are equally capable? \\
\textbf{Label} : Paraphrase \\
\textbf{Category} : {\color{blue}Correct} \\
\\
\textbf{Question 1} : What happened to The Joker after The end of The Dark Knight? \\
\textbf{Question 2} : What happens to the Joker at the end of The Dark Knight (2008 movie)? \\
\textbf{Label} : Non-paraphrase \\
\textbf{Category} : {\color{blue}Wrong} \\
\\
\textbf{Question 1} : I love my wife more then anything. Why do I fantasize about her with other men? \\
\textbf{Question 2} : Why do I fantasize about other men having sex with my wife? \\
\textbf{Label} : Paraphrase \\
\textbf{Category} : Fine \\
\\
\textbf{Question 1} : What is your opinion on the new MacBook Pro Touch Bar? \\
\textbf{Question 2} : What do you think about the OLED touch bar on the new MacBook Pro? \\
\textbf{Label} : Paraphrase \\
\textbf{Category} : {\color{blue}Correct} \\
\\
\textbf{Question 1} : How do I get rid of my negative alter ego? \\
\textbf{Question 2} : How do you get rid of your negative alter ego? \\
\textbf{Label} : Paraphrase \\
\textbf{Category} : {\color{blue}Correct} \\
\\
\textbf{Question 1} : How can I get wifi driver for my hp laptop with windows 7 os? \\
\textbf{Question 2} : How can I get wifi driver for my laptop with windows 7 os? \\
\textbf{Label} : Paraphrase \\
\textbf{Category} : Fine \\
\\
\textbf{Question 1} : What's your attitude towards life? \\
\textbf{Question 2} : What should be your attitude towards life? \\
\textbf{Label} : Paraphrase \\
\textbf{Category} : {\color{blue}Wrong} \\
\\
\textbf{Question 1} : What books would I like if I loved A Song of Ice and Fire? \\
\textbf{Question 2} : Are there books which are similar to A Song of Ice and Fire? \\
\textbf{Label} : Paraphrase \\
\textbf{Category} : Fine \\
\\
\textbf{Question 1} : Why do Muslims think they will conquer the whole world? \\
\textbf{Question 2} : Do you think Muslims will take over the world? \\
\textbf{Label} : Non-paraphrase \\
\textbf{Category} : {\color{blue}Correct} \\
\\
\textbf{Question 1} : Is dark matter a sea of massive dark photons that ripple when galaxy clusters collide and wave in a double slit experiment? \\
\textbf{Question 2} : Does a superfluid dark matter which ripples when Galaxy clusters collide and waves in a double slit experiment relate GR and QM? \\
\textbf{Label} : Paraphrase \\
\textbf{Category} : {\color{blue}Correct} \\
\\
\textbf{Question 1} : What is Batman like? \\
\textbf{Question 2} : What is Batman's personality like? \\
\textbf{Label} : Non-paraphrase \\
\textbf{Category} : {\color{blue}Correct} \\
\\

\subsection{SST-2} \label{sec:decaying-sst-2}

SST-2 is the abbreviation of Stanford Sentiment Treebank \cite{socher-etal-2013-recursive}. 
In this task, the model needs to recognize whether the phrases or sentences reflect positive or negative sentiments.
\\
\\
\textbf{Input} : predictability is the only winner  \\
\textbf{Label} : Negative \\
\textbf{Category} : {\color{blue}Correct} \\
\\
\textbf{Input} : abandon their scripts and go where the moment takes them  \\
\textbf{Label} : Negative \\
\textbf{Category} : {\color{blue}Correct} \\
\\
\textbf{Input} : chases for an hour and then  \\
\textbf{Label} : Positive \\
\textbf{Category} : Unsure \\
\\
\textbf{Input} : provide much more insight than the inside column of a torn book jacket  \\
\textbf{Label} : Negative \\
\textbf{Category} : Unsure \\
\\
\textbf{Input} : a children 's party clown  \\
\textbf{Label} : Negative \\
\textbf{Category} : Fine \\
\\
\textbf{Input} : perhaps even the slc high command found writer-director mitch davis 's wall of kitsch hard going .  \\
\textbf{Label} : Negative \\
\textbf{Category} : {\color{blue}Correct} \\
\\
\textbf{Input} : own placid way  \\
\textbf{Label} : Negative \\
\textbf{Category} : {\color{blue}Correct} \\
\\
\textbf{Input} : get on a board and , uh , shred ,  \\
\textbf{Label} : Negative \\
\textbf{Category} : {\color{blue}Correct} \\
\\
\textbf{Input} : asks what truth can be discerned from non-firsthand experience , and specifically questions cinema 's capability for recording truth .  \\
\textbf{Label} : Positive \\
\textbf{Category} : {\color{blue}Correct} \\
\\
\textbf{Input} : puts the dutiful efforts of more disciplined grade-grubbers  \\
\textbf{Label} : Positive \\
\textbf{Category} : {\color{blue}Correct} \\
\\
\textbf{Input} : filter out the complexity  \\
\textbf{Label} : Positive \\
\textbf{Category} : {\color{blue}Correct} \\
\\
\textbf{Input} : told what actually happened as if it were the third ending of clue  \\
\textbf{Label} : Negative \\
\textbf{Category} : {\color{blue}Correct} \\
\\
\textbf{Input} : is more in love with strangeness than excellence .  \\
\textbf{Label} : Positive \\
\textbf{Category} : {\color{red}Wrong} \\
\\
\textbf{Input} : i found myself howling more than cringing  \\
\textbf{Label} : Positive \\
\textbf{Category} : {\color{blue}Correct} \\
\\
\textbf{Input} : goldbacher draws on an elegant visual sense and a talent for easy , seductive pacing ... but she and writing partner laurence coriat do n't manage an equally assured narrative coinage  \\
\textbf{Label} : Positive \\
\textbf{Category} : Unsure \\
\\
\textbf{Input} : for a thirteen-year-old 's book report  \\
\textbf{Label} : Negative \\
\textbf{Category} : {\color{blue}Correct} \\
\\
\textbf{Input} : a problem hollywood too long has ignored  \\
\textbf{Label} : Negative \\
\textbf{Category} : {\color{blue}Correct} \\
\\
\textbf{Input} : twisted sense  \\
\textbf{Label} : Negative \\
\textbf{Category} : {\color{blue}Correct} \\
\\
\textbf{Input} : a stab at soccer hooliganism  \\
\textbf{Label} : Negative \\
\textbf{Category} : {\color{blue}Correct} \\
\\
\textbf{Input} : sinuously plotted  \\
\textbf{Label} : Negative \\
\textbf{Category} : {\color{blue}Correct} \\
\\
\textbf{Input} : shiner can certainly go the distance , but is n't world championship material  \\
\textbf{Label} : Positive \\
\textbf{Category} : {\color{blue}Correct} \\
\\
\textbf{Input} : holding equilibrium up  \\
\textbf{Label} : Negative \\
\textbf{Category} : {\color{red}Wrong} \\
\\
\textbf{Input} : i am highly amused by the idea that we have come to a point in society where it has been deemed important enough to make a film in which someone has to be hired to portray richard dawson .  \\
\textbf{Label} : Positive \\
\textbf{Category} : {\color{red}Wrong} \\
\\
\textbf{Input} : waters  \\
\textbf{Label} : Negative \\
\textbf{Category} : {\color{red}Wrong} \\
\\
\textbf{Input} : what might have emerged as hilarious lunacy in the hands of woody allen or  \\
\textbf{Label} : Positive \\
\textbf{Category} : {\color{blue}Correct} \\
\\
\textbf{Input} : of those airy cinematic bon bons whose aims -- and by extension , accomplishments -- seem deceptively slight on the surface  \\
\textbf{Label} : Positive \\
\textbf{Category} : {\color{blue}Correct} \\
\\
\textbf{Input} : do n't blame eddie murphy but  \\
\textbf{Label} : Negative \\
\textbf{Category} : {\color{blue}Correct} \\
\\
\textbf{Input} : melodramatic paranormal romance  \\
\textbf{Label} : Negative \\
\textbf{Category} : {\color{blue}Correct} \\
\\
\textbf{Input} : could possibly be more contemptuous of the single female population  \\
\textbf{Label} : Negative \\
\textbf{Category} : {\color{blue}Correct} \\
\\
\textbf{Input} : cremaster 3 '' should come with the warning `` for serious film buffs only ! ''  \\
\textbf{Label} : Negative \\
\textbf{Category} : {\color{blue}Correct} \\
\\
\textbf{Input} : softheaded metaphysical claptrap  \\
\textbf{Label} : Negative \\
\textbf{Category} : {\color{blue}Correct} \\
\\
\textbf{Input} : owed to benigni  \\
\textbf{Label} : Negative \\
\textbf{Category} : Unsure \\
\\
\textbf{Input} : to be a suspenseful horror movie or a weepy melodrama  \\
\textbf{Label} : Positive \\
\textbf{Category} : {\color{blue}Correct} \\
\\
\textbf{Input} : genuinely unnerving .  \\
\textbf{Label} : Positive \\
\textbf{Category} : {\color{blue}Correct} \\
\\
\textbf{Input} : gaping enough to pilot an entire olympic swim team through  \\
\textbf{Label} : Negative \\
\textbf{Category} : {\color{blue}Correct} \\
\\
\textbf{Input} : this is popcorn movie fun with equal doses of action , cheese , ham and cheek ( as well as a serious debt to the road warrior ) , but it feels like unrealized potential  \\
\textbf{Label} : Positive \\
\textbf{Category} : Fine \\
\\
\textbf{Input} : feeling like it was worth your seven bucks , even though it does turn out to be a bit of a cheat in the end  \\
\textbf{Label} : Negative \\
\textbf{Category} : {\color{blue}Correct} \\
\\
\textbf{Input} : pull it back  \\
\textbf{Label} : Negative \\
\textbf{Category} : {\color{blue}Correct} \\
\\
\textbf{Input} : , this is more appetizing than a side dish of asparagus .  \\
\textbf{Label} : Negative \\
\textbf{Category} : {\color{blue}Correct} \\
\\
\textbf{Input} : crime drama  \\
\textbf{Label} : Negative \\
\textbf{Category} : Unsure \\
\\
\textbf{Input} : like most movies about the pitfalls of bad behavior  \\
\textbf{Label} : Negative \\
\textbf{Category} : Fine \\
\\
\textbf{Input} : befallen every other carmen before her  \\
\textbf{Label} : Positive \\
\textbf{Category} : Unsure \\
\\
\textbf{Input} : appeal to those without much interest in the elizabethans ( as well as rank frustration from those in the know about rubbo 's dumbed-down tactics )  \\
\textbf{Label} : Negative \\
\textbf{Category} : Unsure \\
\\
\textbf{Input} : about existential suffering  \\
\textbf{Label} : Negative \\
\textbf{Category} : Fine \\
\\
\textbf{Input} : , if uneven ,  \\
\textbf{Label} : Negative \\
\textbf{Category} : Unsure \\
\\
\textbf{Input} : succumbs to sensationalism  \\
\textbf{Label} : Positive \\
\textbf{Category} : {\color{red}Wrong} \\
\\
\textbf{Input} : that turns me into that annoying specimen of humanity that i usually dread encountering the most  \\
\textbf{Label} : Negative \\
\textbf{Category} : Fine \\
\\
\textbf{Input} : at least a minimal appreciation  \\
\textbf{Label} : Positive \\
\textbf{Category} : Unsure \\
\\
\textbf{Input} : underlines even the dullest tangents  \\
\textbf{Label} : Negative \\
\textbf{Category} : {\color{blue}Correct} \\
\\
\textbf{Input} : heard before  \\
\textbf{Label} : Positive \\
\textbf{Category} : Unsure \\
\\
\textbf{Input} : i like my christmas movies with more elves and snow and less pimps and ho 's .  \\
\textbf{Label} : Negative \\
\textbf{Category} : Unsure \\
\\
\textbf{Input} : can aspire but none can equal  \\
\textbf{Label} : Negative \\
\textbf{Category} : Unsure \\
\\
\textbf{Input} : fathom  \\
\textbf{Label} : Negative \\
\textbf{Category} : Unsure \\
\\
\textbf{Input} : attempt to bring cohesion to pamela 's emotional roller coaster life  \\
\textbf{Label} : Negative \\
\textbf{Category} : Unsure \\
\\
\textbf{Input} : movie version  \\
\textbf{Label} : Positive \\
\textbf{Category} : {\color{red}Wrong} \\
\\
\textbf{Input} : of spontaneity in its execution and a dearth of real poignancy  \\
\textbf{Label} : Positive \\
\textbf{Category} : {\color{blue}Correct} \\
\\

\end{document}